\documentclass[10pt,twocolumn,letterpaper]{article}

\usepackage[pagenumbers]{cvpr} % To force page numbers, e.g. for an arXiv version

\makeatletter
\renewcommand\paragraph{\@startsection{paragraph}{4}{\z@}{1.5ex \@plus .5ex \@minus .2ex}{-1em}{\normalfont\normalsize\bfseries}}
\makeatother

\makeatletter
\g@addto@macro\normalsize{%
  \setlength\abovedisplayskip{6pt plus 2pt minus 2pt}\setlength\belowdisplayskip{6pt plus 2pt minus 2pt}%
  \setlength\abovedisplayshortskip{3pt plus 2pt}\setlength\belowdisplayshortskip{3pt plus 2pt minus 1pt}}
\makeatother



\newcommand{\bench}{TACO-Bench\xspace}
\newcommand{\method}{TACO-Zero\xspace}
\usepackage{xstring}
\makeatletter
\newcommand*{\suppletter}[2]{\expandafter\def\csname supp@#1\endcsname{#2}}
\suppletter{sec:app:analysis}{A}\suppletter{sec:app:impl}{B}\suppletter{sec:app:bench}{C}\suppletter{sec:app:results}{D}\suppletter{sec:app:limits}{E}
\DeclareRobustCommand{\supp}[1]{Supp.~\ifcsname supp@#1\endcsname\textcolor{cvprblue}{\csname supp@#1\endcsname}\else\textbf{\textcolor{red}{??}}\fi}
\makeatother
\usepackage{xspace}
\newcommand{\cmark}{\textcolor{ForestGreen}{\ding{51}}}
\newcommand{\xmark}{\textcolor{BrickRed}{\ding{55}}}
\usepackage{pifont}
\usepackage{multirow}%
\usepackage{colortbl}%
\definecolor{oursrow}{HTML}{EAF6F0}%
\definecolor{laddrow}{HTML}{FDF3E7}%
\definecolor{baserow}{HTML}{EEF3FC}%
\definecolor{succB}{HTML}{D7E3F8}\definecolor{succL}{HTML}{F9DFBE}\definecolor{succO}{HTML}{CBEBDA}\definecolor{succN}{HTML}{ECECEC}%

\newif\ifchk \chktrue
\ifdefined\nolabels\chkfalse\fi
\makeatletter
\DeclareRobustCommand{\chk}[1]{\ifchk\hspace{0pt}\chk@list{#1}\fi}
\def\chk@list#1{\begingroup\color{red}\scriptsize\def\chk@sep{[chk:}%
  \@for\chk@i:=#1\do{\textsuperscript{\chk@sep\chk@i}\def\chk@sep{,}\hspace{0pt}}\textsuperscript{]}\endgroup}
\makeatother

\newcommand{\supptabfont}{\fontsize{8pt}{9.5pt}\selectfont}%
\usepackage{framed}
\definecolor{redraftbg}{RGB}{253,226,226}
\newcommand{\mt}[1]{\texttt{#1}}
\usepackage{placeins}%
\usepackage{afterpage}%

\definecolor{cvprblue}{rgb}{0.21,0.49,0.74}
\usepackage[pagebackref,breaklinks,colorlinks,allcolors=cvprblue]{hyperref}
\newcommand{\webpage}{\href{https://kaist-viclab.github.io/taco-zero-site/}{project page}}

\providecommand{\showmain}{1}
\providecommand{\showsupp}{1}
\newif\ifsuppstarted
\makeatletter
\ifnum\showmain=0
  \AddToHook{shipout/before}{\ifsuppstarted\else\DiscardShipoutBox\fi}
  \let\TZ@protectedwrite\protected@write
  \long\def\protected@write#1#2#3{\begingroup#2\let\protect\@unexpandable@protect\edef\reserved@a{\immediate\write#1{#3}}\reserved@a\endgroup\if@nobreak\ifvmode\nobreak\fi\fi}
\fi
\makeatother
\def\confName{CVPR}
\def\confYear{2027}

\title{\raisebox{-0.02em}{\includegraphics[height=0.734em]{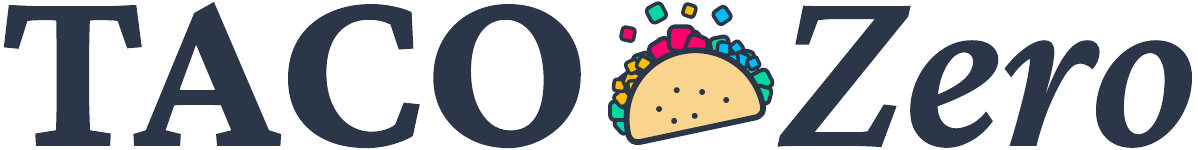}}: Zero-Shot Multi-Reference Compositional Video Editing}
\hypersetup{pdftitle={TACO-Zero: Zero-Shot Multi-Reference Compositional Video Editing}}

\author{%
Samuel Teodoro$^{1}$ \quad Soo Ye Kim$^{2}$ \quad Jihyong Oh$^{3,\dagger}$ \quad Munchurl Kim$^{1,\dagger}$\\[4pt]
$^{1}$VICLab, KAIST \quad $^{2}$Adobe Research \quad $^{3}$CMLab, Chung-Ang University\\
{\tt\small \{sateodoro, mkimee\}@kaist.ac.kr \quad sooyek@adobe.com \quad jihyongoh@cau.ac.kr}\\[4pt]
{\normalsize\url{https://kaist-viclab.github.io/taco-zero-site/}}
}

\begin{document}
\twocolumn[{%
\renewcommand\twocolumn[1][]{#1}%
\maketitle
\begin{center}
  \includegraphics[width=\textwidth]{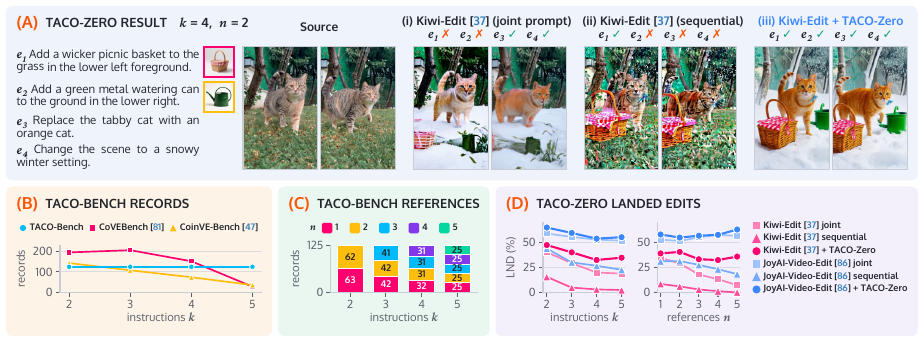}
  \vspace{-6mm}
  \captionof{figure}{\textbf{\method and \bench.} (A)~Given $k = 4$ instructions and $n = 2$ references, the frozen
  Kiwi-Edit~\cite{kiwiedit} lands only the two unreferenced edits from a joint prompt in~(i), and only the first edit when
  applied sequentially in~(ii). With \method in~(iii), it lands all four in one pass, without training. (B)~Other benchmarks~\cite{covebench,coinve} thin out as $k$ grows, while \bench keeps 125 records per $k$, (C)~split evenly over $n = 1$ to $k$.
  (D)~\method lands the most edits on Kiwi-Edit and JoyAI-Video-Edit~\cite{joyai} as $k$ and $n$ grow.}
  \label{fig:teaser}
\end{center}
\vspace{1em}
}]
{\renewcommand{\thefootnote}{\fnsymbol{footnote}}\footnotetext[2]{Co-corresponding authors.}}
\begin{abstract}
Instruction-based video editing changes a clip with words, and
reference images show what words cannot describe. Moreover, users often ask
for several edits at once, some with references. Existing editors
drop or entangle instructions and their references, and training them to
compose edits means rebuilding compositional data for every new task. We instead
compose edits zero-shot with a frozen single-instruction editor. Our
cross-attention analysis shows: (i) each instruction alone finds its target and
lands its edit, (ii) a joint prompt locates every target but entangles the
instructions, and (iii) separate contexts keep the edits apart but weaken them, since
each region splits its attention over all contexts. Building on these, we propose \method (Token Assignment for COmpositional video
editing), which encodes each instruction separately and finds its target region from
cross-attention over a few denoising steps. It then renders all edits at
once, assigning each region's video tokens to attend only to its own
instruction and reference. \method plugs into cross-attention DiTs and
MMDiTs without any additional training. Since existing benchmarks lack references or thin
out as the number of instructions grows, we build \bench, a benchmark with 500
records balanced over two to five instructions and one to five
references, and each instruction scored on its own.
\method lifts the share of landed edits from 24.51\% to 37.20\% on Kiwi-Edit and from 54.11\% to 57.37\% on JoyAI-Video-Edit.\par
\end{abstract}

\section{Introduction}
\label{sec:intro}

Instruction-based video editing lets users modify a clip by describing
their desired result in words. This works well for edits such as a color or weather change.
However, it struggles when the edit must place a particular object in the clip, as text often fails to capture the specific details of an
edit~\cite{dreamomni2}. For example, a user may want their own dog inserted on a
couch,
but text alone cannot reproduce the same dog. Describing an object in detail is also
cumbersome for many users. Reference images avoid both problems, and
recent video editors~\cite{anyv2v,stablev2v,unic,univideo,instructx,easyv2v,viva,kiwiedit,mive,loomvideo,tide,goku,rebind,bernini,omnixref,vino,joyai,omniweaving} have started to support them.

Furthermore, users often ask for multiple edits at once, with some needing a
reference. For example, in \mbox{\cref{fig:teaser}-(A)}, a user may want to
add a picnic basket and a green watering can, each from its own photo, then
turn the cat orange and the scene wintry. Compositional editors~\cite{goku,coinve} do not combine
references with multiple instructions. Single-reference
editors~\cite{kiwiedit,joyai}, trained with one reference per edit, may drop an
instruction or misplace a reference when all instructions and
references are given in a joint prompt (\mbox{\cref{fig:teaser}-(A)(i)}).
Editing one instruction at a time instead accumulates artifacts
(\mbox{\cref{fig:teaser}-(A)(ii)}). Multi-reference
editors~\cite{univideo,loomvideo,vino,bernini,omnixref,rebind,tide} combine
multiple references, but their compositional training data must be rebuilt for every new task, such as scene text editing~\cite{steervte,vitexbench}.

A zero-shot route avoids this cost. The editor is trained on single
instructions only, so a new task needs only its own single-instruction data,
and multiple edits are composed at inference without training. To the best of our
knowledge, no zero-shot method composes multiple referenced edits at once on a frozen
single-instruction video editor.

\begin{figure*}[!t]
  \centering
  \includegraphics[width=\textwidth]{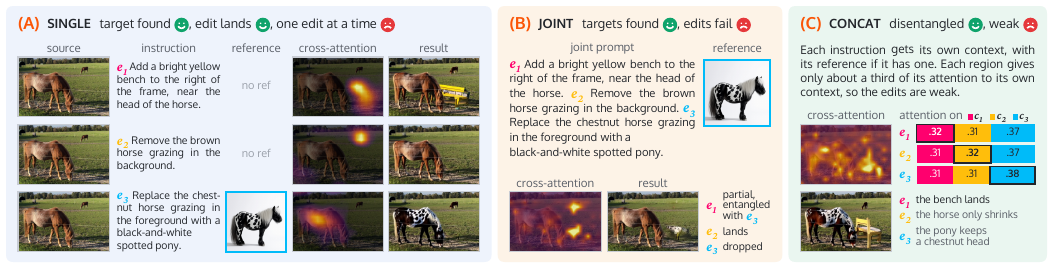}
  \vspace{-6.5mm}
  \caption{\textbf{Analysis of a frozen editor~\cite{kiwiedit} on multiple instructions} (cross-attention at step 5). (A)~Each instruction alone
  locates its target, and its edit lands. (B)~A joint prompt still locates every target, but the instructions share one context,
  so the edits are entangled or dropped. (C)~Concatenated separate contexts keep the edits apart, but each region splits its
  attention over all contexts, so the edits are weak.}
  \label{fig:analysis}
  \label{fig:analysis-leak}
  \label{fig:analysis-loc}
  \label{fig:analysis-share}
  \label{fig:analysis-ref}
\end{figure*}

Existing benchmarks also fall short of this setting. Reference
benchmarks~\cite{kiwiedit,omnieditbench,goku,instructx,unic,editverse,bernini,omnixref,omniweaving}
ask for one edit per sample. Compositional
benchmarks~\cite{covebench,coinve} give multiple instructions per video but no
references, and their records thin out as the number of instructions grows
(\mbox{\cref{fig:teaser}-(B)}).
Multi-reference benchmarks~\cite{univbench,tide} write all instructions and
at most three references into one compound prompt. Thus, how an editor copes as
instructions and references grow cannot be measured reliably.

These problems motivate a zero-shot method that uses a frozen editor's
single-instruction capability to compose multiple edits at inference. We
first analyze the cross-attention of Kiwi-Edit~\cite{kiwiedit}. A single instruction locates its target and lands
its edit, but only one at a time (\mbox{\cref{fig:analysis}-(A)}). A joint prompt still locates every target, but its
instructions share one context, so their edits are entangled or dropped (\mbox{\cref{fig:analysis}-(B)}). Concatenated separate
contexts disentangle the edits but weaken them, since each region splits its attention over all
contexts (\mbox{\cref{fig:analysis}-(C)}). Our method, \textbf{\method} (\underline{T}oken
\underline{A}ssignment for \underline{CO}mpositional video editing), builds on
these findings. It encodes each instruction separately and finds its target region from
cross-attention over a few denoising steps. It then renders all edits at once,
assigning each region's video tokens to its own instruction and
reference.
To evaluate \method, we introduce
\textbf{\bench} with 500 records, 125 for each of two to five instructions (\mbox{\cref{fig:teaser}-(B)}), where one to all
instructions have a reference
(\mbox{\cref{fig:teaser}-(C)}).
To demonstrate generalizability, we plug \method into two editors with
different architectures, the cross-attention DiT Kiwi-Edit~\cite{kiwiedit} and the
MMDiT JoyAI-Video-Edit~\cite{joyai}, improving both backbones' performance. Our contributions are as follows:
\begin{itemize}
  \item An analysis of how a frozen video editor attends to
    multiple instructions, showing why its edits fail to land.
  \item \textbf{\method}, to the best of our knowledge, the first zero-shot method that extends single-instruction video
    editors, both cross-attention DiTs and MMDiTs, to multi-reference compositional video editing.
  \item \textbf{\bench}, a 500-record benchmark balanced over two to five instructions and
    one to five references, with every instruction scored on its own.
  \item Experiments showing that \method lifts landed edits from 24.51\% to
    37.20\% on Kiwi-Edit and from 54.11\% to 57.37\% on JoyAI-Video-Edit.
\end{itemize}

\section{Related Work}
\label{sec:related}

\paragraph{Instruction-based editing.}
Instruction-based editing changes visual content following a natural
language instruction. In image editing, InstructPix2Pix~\cite{ip2p} started
this line of work,
followed by datasets~\cite{magicbrush,hqedit,ultraedit,anyedit} and
editors~\cite{emuedit,omniedit,step1xedit}. Without paired video data,
some video editors run an image model on every
frame~\cite{t2vzero,tokenflow}, on the first frame with image-to-video
propagation~\cite{anyv2v,i2vedit,stablev2v}, or on frame chunks with
cross-frame memory~\cite{editvid}. Recent video
editors~\cite{effived,instructvedit,vace,unic,reco,kiwiedit}
train on million-scale video
datasets~\cite{insvie,vivid10m,senorita,ditto,openve3m}, on images and
videos~\cite{univideo,editverse,bernini,loomvideo,joyai,live}, or on image
pairs with little
video~\cite{instructx,dreamve,insedit,imvideoedit,vifeedit,videocof},
often with a multimodal language model~\cite{qwen25vl,qwen3vl} as the instruction
encoder~\cite{kiwiedit,univideo,instructx,goku,coinve,loomvideo,mive,bernini,omnixref,joyai}.
We keep such editors~\cite{kiwiedit,joyai} frozen rather than train our own.

\paragraph{Compositional editing.}
Users often ask for multiple edits at once. Trained
editors~\cite{coinve,goku,mcie} learn composition from multi-instruction data.
Zero-shot image editors keep the edits apart in a branch per edit
group~\cite{paralleledits,mirage}, a layer per object~\cite{layeredit} or a
LoRA per concept~\cite{cds}, or confine each edit to a given
mask~\cite{mdeedit,promptartisan}. Others
modulate~\cite{fusionedit,foi,videograin,primedit,idagedit} or
bind~\cite{breakingpoint,mice,bindedit,disentangling} the attention of a
U-Net~\cite{unet} or a DiT~\cite{dit} inside each instruction's region, given
by a segmenter~\cite{videograin}, the
user~\cite{breakingpoint,mice,bindedit,primedit,idagedit} or the editor
itself~\cite{fusionedit,foi,disentangling}. Single-edit video
editors~\cite{flowdirector,dfvedit,consistedit} also extract a mask from their
own attention. All of them encode the instruction as text alone, and their
masks come from one keyword of the instruction~\cite{foi,flowdirector} or the
mean over all its tokens~\cite{disentangling}. Our backbones~\cite{kiwiedit,joyai}
encode each instruction with an MLLM together with the source video and its
reference. We find each region from the attention of this whole context. To our knowledge, no other zero-shot method composes multiple
instructions on such an editor.

\paragraph{Reference-guided editing.}
A reference image can show what words cannot describe, but it must stay tied to the
instruction that uses it. Most video editors accept one
reference~\cite{anyv2v,stablev2v,unic,instructx,easyv2v,viva,kiwiedit,mive,goku,joyai}.
Multi-reference image and video
editors~\cite{dreamomni2,univideo,rebind,tide,bernini,vino,omnixref,loomvideo,minimaxh3,seedance25}
learn this tie from data that must be rebuilt for every new task, by concatenating the references with the
video~\cite{univideo} or tagging each with an index~\cite{dreamomni2},
tokens~\cite{rebind,vino}, a task embedding~\cite{tide} or rotary
indices~\cite{bernini,omnixref,loomvideo}. Zero-shot methods
strengthen reference attention for a single edit~\cite{refgap} or pass the
reference to a frozen image editor as is~\cite{editvid}. To our
knowledge, no zero-shot method composes referenced edits on a
frozen single-instruction video editor.

\begin{figure*}[!t]
  \centering
  \includegraphics[width=\linewidth]{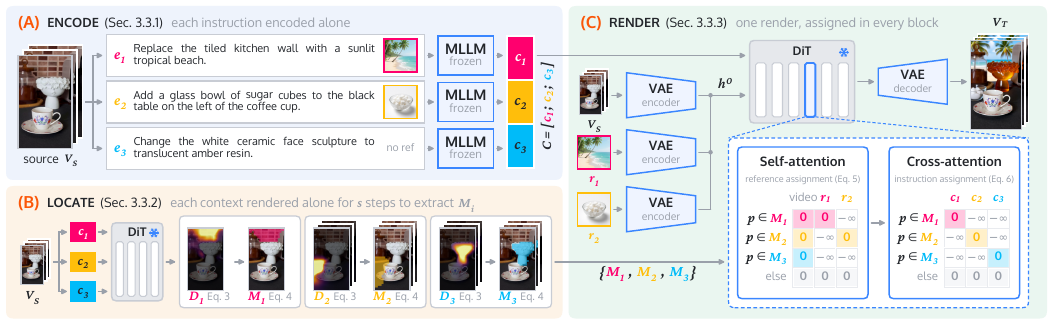}
  \vspace{-7mm}
  \caption{\textbf{\method pipeline.} We run a frozen editor in three stages, here with $k = 3$ and
  $n = 2$. (A)~\emph{Encode.} The editor's MLLM encodes each instruction $e_i$ alone, with its reference if it
  reads one, into its own context $c_i$, and all $c_i$ are concatenated into $C$. (B)~\emph{Locate.} We render each $c_i$ alone
  for $s$ steps, threshold its contrast map $D_i$, and keep in $M_i$ the video tokens where $D_i$ is highest. (C)~\emph{Render.} We render $V_T$ once and, in every
  block, assign each token in $M_i$ to $c_i$ alone in cross-attention and to $r_i$ alone among references in self-attention.}
  \label{fig:pipeline}
\end{figure*}

\paragraph{Video editing benchmarks.}
Most benchmarks~\cite{openve3m,ivebench,videocof,reco} give one instruction
per video and no reference.
Others~\cite{kiwiedit,omnieditbench,goku,instructx,unic,editverse,bernini,omnixref,omniweaving}
add a reference image or video but ask for one edit per sample.
Compositional benchmarks~\cite{covebench,coinve} give multiple edits per
video, but only as text, and recent ones~\cite{tide,univbench} bind multiple
references to the edits of one compound instruction. In all of them, records
thin out as edits or references grow. CoVEBench~\cite{covebench} drops from
206 with three edits to 28 with five, CoinVE-Bench~\cite{coinve} from
143 with two to 36 with five, and UniVBench~\cite{univbench} from 573
with two references to 44 with three, while TIDE-Bench~\cite{tide} caps
references at three. Our \bench holds 125 records for each number of
instructions from two to five (\mbox{\cref{fig:teaser}-(B)}), with one to five
references (\mbox{\cref{fig:teaser}-(C)}), and scores each instruction alone.

\section{Method}
\label{sec:method}

\subsection{Preliminaries}
\label{sec:prelim}

We build on two instruction-based video
editors~\cite{kiwiedit,joyai}, both built on a DiT~\cite{dit} trained with
flow matching~\cite{flowmatching}. Each editor $\mathcal{F}(V_S, e, r)$ edits a source
video $V_S$ by an instruction $e$, with an \textit{optional} reference image $r$. A
multimodal large language model (MLLM)
$\mathcal{M}$~\cite{qwen25vl,mimovl} reads $e$ with $V_S$ and $r$~\cite{kiwiedit},
or with only the first frame of $V_S$~\cite{joyai}, and gives the
context tokens $c$. Given $c$, the DiT
$v_\theta(x_t, t, c)$ regresses the velocity $x_1 - x_0$ along the straight
path~\cite{rectifiedflow} $x_t = t\,x_1 + (1-t)\,x_0$ from Gaussian noise
$x_0$ to the video latent $x_1$, where the time $t \in [0, 1]$ runs from noise to video. The DiT input is
\begin{equation}
  h^{0} = \big[\, P(x_t) \oplus P_S(\mathcal{E}(V_S)) \,;\;
          P_r(\mathcal{E}(r)) \,\big] ,
  \label{eq:input}
\end{equation}
where $\mathcal{E}$ is the VAE encoder, $P$, $P_S$ and $P_r$ are patch
embeddings, $[\,\cdot\,;\,\cdot\,]$ concatenates tokens, and $\oplus$ adds
the scaled source tokens to the noisy ones~\cite{kiwiedit} or
concatenates them~\cite{joyai}. Each DiT block takes the tokens $h$ of the previous block, starting from $h^0$. Each attention layer computes
\begin{equation}
  \mathrm{Attn}(Q, K, V) = \mathrm{softmax}\!\left(QK^{\top}/\sqrt{d}\right) V ,
  \label{eq:xattn}
\end{equation}
where $d$ is the head dimension and $Q$ comes from the video tokens of $h$. In
a cross-attention DiT~\cite{wan,kiwiedit}, $h$ attends to itself
in self-attention and to $c$ in cross-attention. In an
MMDiT~\cite{sd3,joyai}, one joint attention covers $[h; c]$, and we refer to its
video-to-video and video-to-context parts as self-attention and
cross-attention, respectively.

\paragraph{Problem setting.}
Given a source video $V_S$, $k$ atomic instructions $e_1,\dots,e_k$, each asking for one edit, and $n \le k$ reference images, where $r_i$ shows what the edit of $e_i$
should look like for $i \le n$, the goal is a target video $V_T$ in which every edit lands. The edit of $e_i$
\emph{lands} when it is made on its target, looks natural and stays off the other targets. We assume the targets are
disjoint (\supp{sec:app:limits}). We use a frozen editor $\mathcal{F}$ trained on one instruction at a time and add no training.
The common baseline gives $\mathcal{F}$ all $k$ instructions in one \emph{joint prompt}, with all $n$ references.

\subsection{Cross-attention analysis}
\label{sec:analysis}

Given $k > 1$ instructions, a frozen editor $\mathcal{F}$ often fails this test. Some edits are dropped, and others are entangled. To understand what decides whether an edit lands, we study
Kiwi-Edit~\cite{kiwiedit} on a clip with three instructions, one of
them with a reference. We take the cross-attention between the video tokens $h$ and the
context at step 5 of denoising
and compare it with the result in \cref{fig:analysis}. We find
three behaviors, and each one leads to a stage of \method (\cref{sec:ours}).

\paragraph{(A) A single instruction locates its target, and its edit lands.}
When each instruction is encoded alone, its cross-attention already sits on
its target by step 5, and its edit lands (\mbox{\cref{fig:analysis}-(A)}).
A few denoising steps are therefore enough to locate the region of the video
that an instruction affects.

\paragraph{(B) A joint prompt locates every target, but its instructions entangle.}
Given one joint prompt, all instructions share one context. The
cross-attention still covers every target (\mbox{\cref{fig:analysis}-(B)}), so the
editor knows where each edit goes. Yet the bench takes on the pony's black-and-white spots, and
the pony is dropped. The problem is what the target region of each $e_i$ attends to. Of the 128
context tokens a region attends to most, 117 to 120 are shared with other
regions (\supp{sec:app:analysis}). Each region thus attends to a mix of all instructions. An edit then takes
on another instruction's content, as the bench does, or is overridden, as the pony is.

\paragraph{(C) Concatenated contexts keep the edits apart, but they land weakly.}
When each instruction is encoded alone and the contexts are concatenated, each
instruction $e_i$ has its own tokens $c_i$. The cross-attention again covers every target
(\mbox{\cref{fig:analysis}-(C)}), and the edits are no longer entangled. The bench lands,
but the other two are weak. The pony keeps a chestnut head, and the
horse at the back only shrinks. Each region still spreads its attention over
the three contexts, and only 0.32 to 0.38 of that attention goes to its own
$c_i$. Rendered alone, the same region attends only to $c_i$.
Here each edit gets a third of that attention, so the edits are weak.

\subsection{\texorpdfstring{\method}{TACO-Zero}}
\label{sec:ours}

\paragraph{Overview.}
\Cref{sec:analysis} shows that a frozen editor already locates every target,
and that its edits fail because of what each target region attends to. Building on
these findings, we propose \underline{T}oken \underline{A}ssignment for
\underline{CO}mpositional video editing, or \method, a zero-shot method that lets a frozen
single-instruction editor perform multiple edits at once, each with its own
reference if it has one. It runs the
frozen editor~\cite{kiwiedit,joyai} in three stages, \textit{Encode}, \textit{Locate} and \textit{Render}, shown in
\cref{fig:pipeline}. (A)~\emph{Encode}: we encode each $e_i$
alone, with $r_i$ if the MLLM reads one, into its own context $c_i$, and set
$C = [c_1; \dots; c_k]$. (B)~\emph{Locate}: we render each $c_i$ alone for $s$
steps and turn its contrast map $D_i$ into a set $M_i$ of video tokens.
(C)~\emph{Render}: we render $V_T$ once and, in every block, assign each
token in $M_i$ to $r_i$ alone among references in self-attention and to $c_i$
alone in cross-attention.

\subsubsection{Encode}
\label{sec:ours:encode}
A joint prompt drops or entangles edits (\mbox{\cref{fig:analysis}-(B)}), while separately encoded and
concatenated contexts keep them apart (\mbox{\cref{fig:analysis}-(C)}). \textit{Encode} thus gives each $e_i$ its
own context $c_i = \mathcal{M}(V_S, e_i, r_i)$, with $r_i$ only if $e_i$ has one and $\mathcal{M}$ reads images,
and sets $C = [c_1; \dots; c_k]$ (\mbox{\cref{fig:pipeline}-A}). CoinVE-Edit~\cite{coinve} also encodes each $e_i$
separately but is fine-tuned on these contexts, whereas our editor stays frozen. Alone, \textit{Encode} lands fewer
edits (KE + E, \mt{LND} 24.51 $\rightarrow$ 19.94, \cref{tab:main}), as each region splits its attention over all $k$ contexts, but
it gives \textit{Locate} and \textit{Render} one context per $e_i$.

\begin{table*}[t]
  \centering
  \caption{\textbf{Metrics of \bench.} \bench scores 15 metrics in five dimensions. The judge, Qwen3.5-122B-A10B~\cite{qwen35},
  answers checklist questions, each tied to one instruction (\cref{sec:bench:scoring}). If $e_i$ is not
  executed, its \mt{EQ}, \mt{BND} and \mt{RA} score zero. The first three dimensions score each instruction, but \mt{ALL}
  scores each record, and the last two dimensions score each video. \underline{For every metric, higher is better.}}
  \label{tab:metrics}
  \fontsize{7.3pt}{8.8pt}\selectfont\setlength{\tabcolsep}{8.4pt}\renewcommand{\arraystretch}{0.98}
  \begin{tabular}{@{}>{\raggedright\arraybackslash}p{0.6in}lclcl@{}}
    \toprule
    Dimension & Metric & Abbr. & Evaluator & Range & Description \\
    \midrule
    \multirow{3}{*}{\shortstack[l]{Instruction\\compliance}}
      & Execution & \mt{EXE} & Judge~\cite{qwen35} + checklist & [0, 100] & The edit was made, and what it replaced is gone \\
      & Edit quality & \mt{EQ} & Judge~\cite{qwen35} + checklist & [0, 100] & The edit stays in its area, lasts to the end and looks natural \\
      & Binding & \mt{BND} & Judge~\cite{qwen35} + checklist & [0, 100] & The edit does not spill onto another instruction's target \\
    \midrule
    \multirow{2}{*}{\shortstack[l]{Reference\\fidelity}}
      & Reference attributes & \mt{RA} & Judge~\cite{qwen35} + checklist & [0, 100] & The edit matches each checked detail of $r_i$ \\
      & Reference similarity & \mt{RS} & SAM~3~\cite{sam3} + DINOv2~\cite{dinov2} & [0, 1] & Similarity between $r_i$ and the new content SAM~3 cuts out \\
    \midrule
    \multirow{2}{*}{Success}
      & Landing & \mt{LND} & Judge~\cite{qwen35} + checklist & [0, 100] & Every \mt{EXE}, \mt{EQ} and \mt{BND} answer for an instruction is right \\
      & All landed & \mt{ALL} & Judge~\cite{qwen35} + checklist & [0, 100] & Every instruction of a record passes \mt{LND} \\
    \midrule
    \multirow{4}{*}{\shortstack[l]{Video\\fidelity}}
      & Semantic consistency & \mt{SEM} & Judge~\cite{qwen35} rating & [1, 10] & Judge rating of how well untouched things are kept \\
      & Static region consistency & \mt{SRC} & SAM~3~\cite{sam3} + DINOv2~\cite{dinov2} & [0, 1] & Similarity of each untouched thing inside its source mask \\
      & Structure outside edits & \mt{SSO} & SSIM~\cite{ssim} & [0, 1] & SSIM of the pixels outside every edit region \\
      & Motion fidelity & \mt{MF} & CoTracker3~\cite{cotracker3} & [0, 1] & Trajectory similarity between source and result \\
    \midrule
    \multirow{4}{*}{\shortstack[l]{Video\\quality}}
      & Comprehensive quality & \mt{VQR} & VisualQuality-R1~\cite{visualqualityr1} & [1, 5] & Holistic visual quality from a visual reward model \\
      & Aesthetic quality & \mt{AES} & Aesthetic Predictor v2.5~\cite{aesv25} & [1, 10] & Frame-level aesthetic appeal \\
      & Technical quality & \mt{TQ} & DOVER++~\cite{dover} & (0, 100) & Blur, noise, compression, flicker and jitter \\
      & Motion smoothness & \mt{MSM} & RAFT~\cite{raft} & [0, 1] & Consistency of optical flow between adjacent frames \\
    \bottomrule
  \end{tabular}
\end{table*}

\subsubsection{Locate}
\label{sec:ours:locate}
Concatenated contexts land weak edits (\mbox{\cref{fig:analysis}-(C)}), so \textit{Locate} finds the
video tokens that should attend to each $c_i$ by rendering $V_S$ with each $c_i$ alone for $s$ steps
(\mbox{\cref{fig:pipeline}-B}). At step $s$,
$F_i^{l}$ is the cross-attention map of $c_i$ in block $l$, normalized to sum
to 1. We score each block by
$\delta_l = \mathrm{mean}_{i<j}\, \lVert F_i^{l} - F_j^{l} \rVert_1$, which is
high when the contexts attend to different regions and low when they attend to
the same ones. $F_i$ is the $\delta_l$-weighted average of $F_i^{l}$ over the
$L$ blocks with the highest $\delta_l$. We then
subtract the other instructions' mean map, which removes regions they all
attend to, and smooth with a Gaussian $G$ on the latent grid:
\begin{equation}
  D_i = G \ast \big( F_i - \tfrac{1}{k-1} {\textstyle\sum_{j \neq i}} F_j \big) .
  \label{eq:contrast}
\end{equation}
For a video token $p$, $D_i(p)$ is the contrast of $e_i$ at $p$. With $\gamma_i$ the Otsu threshold~\cite{otsu} of $D_i$, the candidate set of $e_i$ is $\tilde{M}_i = \{p : D_i(p) > \gamma_i\}$.
These sets can overlap, so $M_i$ keeps $p \in \tilde{M}_i$ only if $D_i(p)$ is the largest over sets with $p$:
\begin{equation}
  M_i = \big\{\, p \in \tilde{M}_i : D_i(p) \ge D_j(p)
        \;\, \forall j \text{ with } p \in \tilde{M}_j \,\big\} .
  \label{eq:mask}
\end{equation}

\subsubsection{Render}
\label{sec:ours:render}
\paragraph{Reference assignment in self-attention.}
Each reference $r_i$ is appended to $h^{0}$ as its own token set $\mathcal{T}_i$ (\cref{eq:input}), so
every video token attends to every reference, and a reference leaks into other regions. We therefore let a video token $p$
in $M_i$ attend to $r_i$ and to no other reference. For any query
token $p$ and a key token $u$, we add a bias $A^{\mathrm{self}}_{pu}$ to the
self-attention logits,
$\mathrm{softmax}(QK^{\top}/\sqrt{d} + A^{\mathrm{self}})\,V$, with
\begin{equation}
  A^{\mathrm{self}}_{pu} = \begin{cases}
    -\infty & p \in M_i,\ u \in \mathcal{T}_j,\ j \neq i, \\
    -\infty & p \in \mathcal{T}_i,\ u \in \mathcal{T}_j,\ j \neq i, \\
    -\infty & p \in \mathcal{V},\ u \in \mathcal{B}, \\
    0 & \text{otherwise},
  \end{cases}
  \label{eq:route}
\end{equation}
where $\mathcal{V}$ is the set of video tokens and $\mathcal{B}$ is the set of reference tokens on the padded border. A video token in $M_i$ thus
attends to the video and $r_i$, or to the video alone if $e_i$ has none. A video token not in any $M_i$ attends to the video and every
reference. Each reference attends to the video and itself. This raises the preservation of the untouched
parts (\mt{SSO}) from 0.460 to 0.532 (KE + E vs KE + E + R, \cref{tab:main}).

\paragraph{Instruction assignment in cross-attention.}
With concatenated contexts $C$, each region spreads its attention over all $k$ contexts, so its edit lands
weakly (\mbox{\cref{fig:analysis}-(C)}). Instruction assignment fixes this by giving each region its own context alone: inside $M_i$,
we remove every other context (\mbox{\cref{fig:pipeline}-C}) by adding $A^{\mathrm{cross}}_{pu}$ to the cross-attention
logits:
\begin{equation}
  A^{\mathrm{cross}}_{pu} = \begin{cases}
    -\infty & p \in M_i,\ u \in c_j,\ j \neq i, \\
    0 & \text{otherwise}.
  \end{cases}
  \label{eq:xroute}
\end{equation}
A video token in $M_i$ thus attends only to $c_i$, as when rendered alone,
and a video token not in any $M_i$ attends to every $c_i$. This raises landed edits (\mt{LND})
from 19.94\% to 33.60\% (\mbox{KE + E} vs \mbox{KE + E + I}, \cref{tab:main}).

\section{\bench}
\label{sec:bench}

Our setting needs a benchmark that (i) varies $k$ and $n$, and (ii) tells which
instruction failed, but none exists (\supp{sec:app:bench}).
Compositional benchmarks~\cite{covebench,coinve} have no references, and their
records thin out as $k$ grows. Multi-reference benchmarks~\cite{univbench,tide}
write all edits and at most three references into one compound prompt, so
neither $k$ nor $n$ is controlled. Only CoVEBench scores each instruction on its own. We therefore build \bench, with
125 records per $k$ from 2 to 5 (\mbox{\cref{fig:teaser}-(B)}), split
evenly over $n$ (\mbox{\cref{fig:teaser}-(C)}). Each $r_i$ is bound to $e_i$,
and each instruction has its own questions.

\begin{table*}[t]
  \centering
  \caption{\textbf{Quantitative comparison on \bench.} Params: the sizes of $\mathcal{M}$ and the DiT. Input: the most instructions and
  references taken in one pass. Blue rows are joint prompts; orange rows add our per-instruction encoding (E), reference
  assignment (R) or instruction assignment (I); green rows add all three (\method). Shaded: \mt{LND}, the share of instructions whose \mt{EXE}, \mt{EQ} and \mt{BND} answers are all right, and \mt{ALL}, the share of records whose instructions all land. Per backbone, without sequential, \textbf{bold} is best, \underline{underlined} second.}
  \label{tab:main}
  \fontsize{7.2pt}{8.7pt}\selectfont\setlength{\tabcolsep}{3.65pt}
  \begin{tabular}{@{}lcc*{15}{c}@{}}
    \toprule
    \multirow{2}{*}{Method} & \multirow{2}{*}[-0.6ex]{\shortstack[c]{Params (B)\\$\mathcal{M}$ + DiT}} & \multirow{2}{*}{Input} & \multicolumn{3}{c}{Instr.\ compliance} & \multicolumn{2}{c}{Ref.\ fidelity} & \multicolumn{2}{>{\columncolor{succN}}c}{\textbf{Success}} & \multicolumn{4}{c}{Video fidelity} & \multicolumn{4}{c}{Video quality} \\
    \cmidrule(lr){4-6}\cmidrule(lr){7-8}\cmidrule(lr){9-10}\cmidrule(lr){11-14}\cmidrule(l){15-18}
    & & & \mt{EXE} & \mt{EQ} & \mt{BND} & \mt{RA} & \mt{RS} & \cellcolor{succN}\textbf{\mt{LND}} & \cellcolor{succN}\textbf{\mt{ALL}} & \mt{SEM} & \mt{SRC} & \mt{SSO} & \mt{MF} & \mt{VQR} & \mt{AES} & \mt{TQ} & \mt{MSM} \\
    \midrule
    Seedance 2.5~\cite{seedance25} & -- & 1 / $n$ & 87.32 & 75.10 & 76.40 & 77.08 & 0.733 & \cellcolor{succN}57.93 & \cellcolor{succN}23.94 & 8.80 & 0.883 & 0.666 & 0.955 & 3.96 & 4.97 & 15.31 & 0.753 \\
    \midrule
    CoinVE-Edit~\cite{coinve} & 8 + 14 & $k$ / 0 & 87.87 & 76.48 & 76.03 & 59.26 & 0.631 & \cellcolor{succN}61.89 & \cellcolor{succN}27.20 & 9.16 & 0.875 & 0.663 & 0.915 & 4.27 & 5.53 & 17.58 & 0.752 \\
    Bernini~\cite{bernini} & 7 + 14 & 1 / $n$ & 86.56 & 73.71 & 75.65 & 69.33 & 0.701 & \cellcolor{succN}55.89 & \cellcolor{succN}25.40 & 8.84 & 0.896 & 0.755 & 0.956 & 4.22 & 5.32 & 16.70 & 0.760 \\
    MiniMax-H3~\cite{minimaxh3} & 32 + 33 & 1 / $n$ & 77.18 & 71.23 & 77.19 & 70.99 & 0.721 & \cellcolor{succN}56.00 & \cellcolor{succN}17.40 & 9.20 & 0.894 & 0.732 & 0.936 & 4.37 & 5.21 & 17.39 & 0.761 \\
    LoomVideo~\cite{loomvideo} & 8 + 5 & 1 / $n$ & 82.92 & 64.44 & 63.72 & 58.34 & 0.685 & \cellcolor{succN}42.63 & \cellcolor{succN}12.40 & 7.08 & 0.816 & 0.582 & 0.917 & 4.26 & 5.31 & 16.06 & 0.753 \\
    Kiwi-Edit-Ref~\cite{kiwiedit} & 3 + 5 & 1 / $n$ & 50.60 & 43.00 & 29.07 & 31.35 & 0.627 & \cellcolor{succN}22.91 & \cellcolor{succN}5.00 & 8.51 & 0.860 & 0.714 & 0.941 & 4.01 & 4.93 & 15.83 & 0.758 \\
    ReCo~\cite{reco} & -- + 1 & 1 / 1 & 65.24 & 27.26 & 38.21 & 43.81 & 0.672 & \cellcolor{succN}8.17 & \cellcolor{succN}1.00 & 5.33 & 0.712 & 0.296 & 0.899 & 2.04 & 4.22 & 11.60 & 0.760 \\
    LTX-Ripple~\cite{ltxripple} & 12 + 19 & 1 / $n$ & 92.67 & 76.41 & 82.68 & 82.42 & 0.720 & \cellcolor{succN}56.80 & \cellcolor{succN}24.60 & 7.98 & 0.841 & 0.553 & 0.932 & 4.26 & 5.39 & 16.27 & 0.743 \\
    EditVid~\cite{editvid} & -- + 9 & 1 / $n$ & 82.65 & 67.30 & 72.95 & 68.46 & 0.679 & \cellcolor{succN}50.97 & \cellcolor{succN}18.20 & 9.14 & 0.917 & 0.825 & 0.938 & 4.20 & 5.45 & 16.20 & 0.714 \\
    \midrule
    Kiwi-Edit (KE)~\cite{kiwiedit}, sequential & 3 + 5 & 1 / 1 & 51.54 & 20.29 & 24.16 & 32.46 & 0.621 & \cellcolor{succN}4.97 & \cellcolor{succN}1.00 & 5.91 & 0.731 & 0.292 & 0.919 & 2.25 & 4.45 & 12.84 & 0.706 \\
    \rowcolor{baserow}
    KE, joint prompt & 3 + 5 & 1 / $n$ & 61.27 & 42.68 & 34.46 & 32.27 & \underline{0.617} & \cellcolor{succB}24.51 & \cellcolor{succB}5.40 & 7.11 & 0.794 & \underline{0.531} & 0.910 & 3.86 & 4.80 & 15.12 & \textbf{0.756} \\
    \rowcolor{laddrow}
    KE + E & 3 + 5 & $k$ / $n$ & 55.62 & 38.06 & 32.44 & 27.16 & 0.592 & \cellcolor{succL}19.94 & \cellcolor{succL}4.80 & 7.29 & 0.792 & 0.460 & 0.922 & 3.97 & 4.92 & 14.98 & 0.753 \\
    \rowcolor{laddrow}
    KE + E + R & 3 + 5 & $k$ / $n$ & 53.68 & 41.88 & 33.59 & 28.78 & 0.595 & \cellcolor{succL}23.26 & \cellcolor{succL}5.40 & \textbf{8.12} & \textbf{0.812} & \textbf{0.532} & \textbf{0.939} & \underline{4.10} & \textbf{4.98} & \underline{15.94} & \underline{0.755} \\
    \rowcolor{laddrow}
    KE + E + I & 3 + 5 & $k$ / $n$ & \textbf{75.65} & \underline{53.59} & \underline{58.81} & \textbf{47.54} & \underline{0.617} & \cellcolor{succL}\underline{33.60} & \cellcolor{succL}\underline{7.60} & 6.99 & 0.784 & 0.438 & 0.921 & 4.02 & 4.90 & 15.04 & 0.753 \\
    \rowcolor{oursrow}
    KE + \textbf{\method} (Ours) & 3 + 5 & $k$ / $n$ & \underline{74.60} & \textbf{56.27} & \textbf{59.77} & \underline{47.26} & \textbf{0.619} & \cellcolor{succO}\textbf{37.20} & \cellcolor{succO}\textbf{10.00} & \underline{7.86} & \underline{0.805} & 0.505 & \underline{0.937} & \textbf{4.11} & \underline{4.96} & \textbf{15.98} & 0.753 \\
    \midrule
    JoyAI (JAI)~\cite{joyai}, sequential & 7 + 16 & 1 / 1 & 92.49 & 59.24 & 80.85 & 74.67 & 0.644 & \cellcolor{succN}28.51 & \cellcolor{succN}7.40 & 7.71 & 0.767 & 0.553 & 0.926 & 2.26 & 4.29 & 10.61 & 0.726 \\
    \rowcolor{baserow}
    JAI, joint prompt & 7 + 16 & 1 / $n$ & 87.26 & 72.96 & 77.38 & 64.99 & 0.635 & \cellcolor{succB}54.11 & \cellcolor{succB}20.80 & 9.26 & 0.838 & 0.740 & 0.946 & \underline{3.65} & \textbf{4.44} & \textbf{13.81} & 0.742 \\
    \rowcolor{laddrow}
    JAI + E & 7 + 16 & $k$ / $n$ & 66.96 & 61.41 & 55.15 & 44.78 & 0.608 & \cellcolor{succL}45.03 & \cellcolor{succL}15.80 & \textbf{9.51} & \underline{0.847} & \textbf{0.834} & \textbf{0.955} & 3.60 & \underline{4.41} & 13.62 & \textbf{0.748} \\
    \rowcolor{laddrow}
    JAI + E + R & 7 + 16 & $k$ / $n$ & 67.51 & 60.93 & 54.38 & 44.42 & 0.607 & \cellcolor{succL}44.69 & \cellcolor{succL}15.80 & \underline{9.48} & \textbf{0.848} & \underline{0.833} & \underline{0.954} & 3.60 & 4.40 & 13.60 & \underline{0.747} \\
    \rowcolor{laddrow}
    JAI + E + I & 7 + 16 & $k$ / $n$ & \underline{87.45} & \underline{75.01} & \underline{80.65} & \underline{65.89} & \underline{0.637} & \cellcolor{succL}\textbf{57.77} & \cellcolor{succL}\underline{22.80} & 9.38 & 0.841 & 0.784 & 0.950 & \textbf{3.66} & 4.40 & \underline{13.74} & \underline{0.747} \\
    \rowcolor{oursrow}
    JAI + \textbf{\method} (Ours) & 7 + 16 & $k$ / $n$ & \textbf{88.15} & \textbf{75.18} & \textbf{81.23} & \textbf{66.69} & \textbf{0.638} & \cellcolor{succO}\underline{57.37} & \cellcolor{succO}\textbf{23.00} & 9.37 & 0.841 & 0.783 & 0.949 & \textbf{3.66} & 4.39 & 13.71 & \underline{0.747} \\
    \bottomrule
  \end{tabular}
\end{table*}

\subsection{Construction}
\label{sec:bench:records}
\label{sec:bench:construction}

\paragraph{Videos.}
\label{sec:bench:videos}
We fetch 1080p clips of at least ten seconds from Pexels~\cite{pexels}, without shot cuts or watermarks, and keep a clip if
SAM~3~\cite{sam3} tracks at least two disjoint objects that stay visible. Among clips that pass a quality
bar~\cite{dover,aesv25}, we select 330 and check each by hand. Of these, 160 serve one record and 170 serve two, giving 500 records;
two records on the same clip share neither an instruction nor a reference.

\paragraph{Instructions.}
\label{sec:bench:instructions}
For each clip, Qwen3-VL-8B~\cite{qwen3vl} writes an ordered pool of
instructions, one per target that SAM~3 found, so targets stay disjoint. The
pool spans eight edit tasks and puts the three
reference tasks (insertion, swap and background replacement) first. A record
takes the first $k$ instructions of the pool, $e_1,\dots,e_k$, and the first
$n$ of them carry a reference. An instruction with a reference names its
object only by class, color and material, leaving the rest of its appearance to the reference; one without a reference
describes it in words. All $1{,}750$ instructions are used once.

\paragraph{References.}
\label{sec:bench:references}
Qwen3-VL-8B also writes a detailed description of each $e_i$, used to
generate $r_i$. From the description, Qwen-Image~\cite{qwenimage} generates four candidate
images, and we keep the one SigLIP~\cite{siglip} ranks best. Qwen3-VL-8B
then checks that every attribute in the description is visible. Each of the
$1{,}122$ references belongs to one record.

\begin{figure*}[t]
  \centering
  \includegraphics[width=\linewidth]{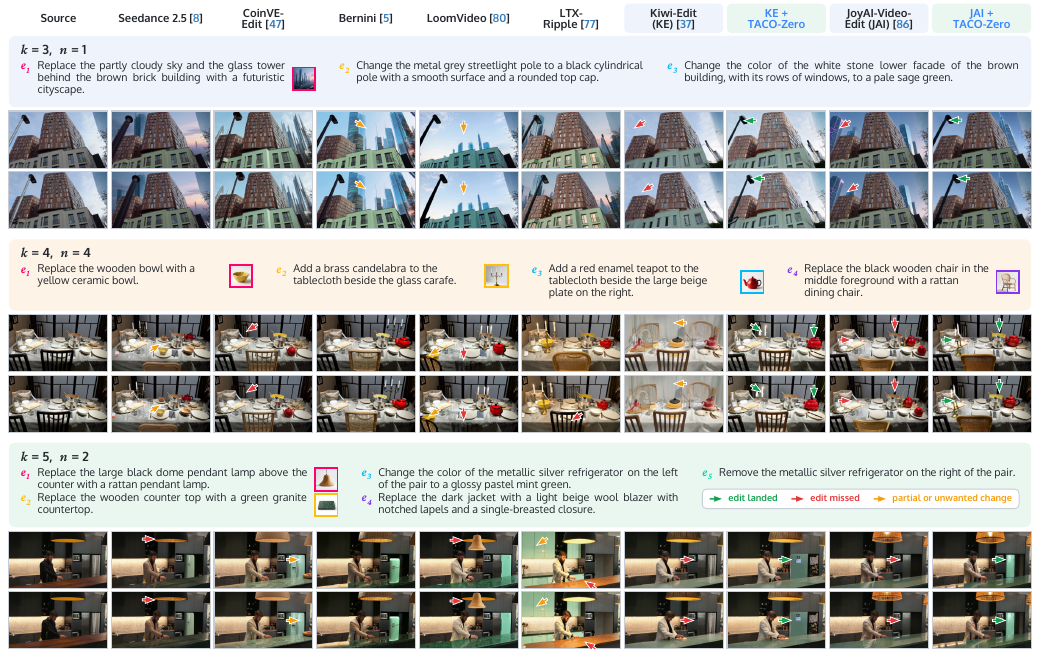}
  \vspace{-7mm}
  \caption{\textbf{Qualitative comparison on \bench.} Columns follow the order of \cref{tab:main}, and each of our
  backbones, Kiwi-Edit~\cite{kiwiedit} and JoyAI-Video-Edit~\cite{joyai}, is followed by \method on it. Videos of more records are on our \webpage.}
  \label{fig:qual}
\end{figure*}

\subsection{Evaluation}
\label{sec:bench:scoring}

\paragraph{Questions.}
Following CoVEBench~\cite{covebench} and
CoinVE-Bench~\cite{coinve}, we score edits with yes/no and
A/B checklist questions, each tied to one instruction. For each $e_i$,
GLM-4.5V~\cite{glm45v} reads $V_S$, $e_i$ and $r_i$ and names the old and new
object, an untouched neighbor and each attribute of $r_i$. Templates turn
these into the questions of \cref{tab:metrics}: was the edit made
(\mt{EXE}), does the neighbor stay, does it last and look natural (\mt{EQ}), does it
leak onto another target (\mt{BND}), and does it show each reference attribute (\mt{RA}).

\paragraph{Judge.}
As in CoVEBench~\cite{covebench}, the judge, Qwen3.5-122B-A10B~\cite{qwen35}, differs from
GLM-4.5V, which writes the questions, so that no model grades its own questions. Execution and absence questions ask whether something in the source changed,
so for them the judge sees both the source and the result. For every other
question, the judge sees the result alone. When an execution question of $e_i$ fails, we
set its persistence, realism, binding and reference scores to zero.
CoVEBench and CoinVE-Bench count such questions even when the edit was never
made, so a skipped edit still earns every ``no leak'' and ``no artifact'' point.

\paragraph{Agreement with human evaluators.}
We check the judge against 27 people, three per video, on 180 edited videos, counting
matching raw answers as in CoVEBench~\cite{covebench}. The judge gives the same answer
as a single rater on 81.5\% of the questions (Cohen's $\kappa = 0.63$) and as the majority
of the three raters on 85.2\% ($\kappa = 0.70$). A $\kappa$ above 0.6 marks substantial
agreement~\cite{landis1977}, so the judge's scores align with human judgment.
We will release the benchmark. More benchmark details and sample records are in \supp{sec:app:bench}.

\section{Experiments}
\label{sec:exp}

\label{sec:exp:setup}

\paragraph{Benchmarks.}
We evaluate on our \bench (\cref{sec:bench}) and report results on another
benchmark in \supp{sec:app:results}.

\paragraph{Baselines.}
We compare with eleven editors (\cref{tab:main}). Most take one prompt, so
they get the joint prompt and all $n$ references: the commercial
Seedance~2.5~\cite{seedance25}, the multi-reference editors
Bernini~\cite{bernini} and LoomVideo~\cite{loomvideo}, the video generator
MiniMax-H3~\cite{minimaxh3}, and the single-reference editors
Kiwi-Edit~\cite{kiwiedit}, its reference-only version Kiwi-Edit-Ref and
JoyAI-Video-Edit~\cite{joyai}. Kiwi-Edit reads each reference after a label
``Reference image $j$'', and JoyAI-Video-Edit encodes each alone and joins
them in its attention cache. LTX-Ripple~\cite{ltxripple} propagates a first
frame that the image editor FLUX.2-Klein~\cite{flux2} edits from the joint
prompt and references, and the zero-shot EditVid~\cite{editvid} runs
FLUX.2-Klein on the video. The compositional editor
CoinVE-Edit~\cite{coinve} reads each instruction separately but takes no
reference. ReCo~\cite{reco} takes one instruction and one reference, so we run
it sequentially, one instruction per pass. We also run Kiwi-Edit and
JoyAI-Video-Edit this way. We skip
UniVideo~\cite{univideo}, VINO~\cite{vino} and OmniWeaving~\cite{omniweaving}:
the license of their base, HunyuanVideo~\cite{hunyuanvideo}, excludes our territory.

\paragraph{Metrics.}
Beside the human-aligned judge-based metrics of \cref{sec:bench:scoring}, we report two metrics
derived from them (\mt{LND} and \mt{ALL}) and metrics from specialized evaluators. \Cref{tab:metrics}
summarizes and describes all of them.

\begin{figure*}[t]
  \centering
  \includegraphics[width=\textwidth]{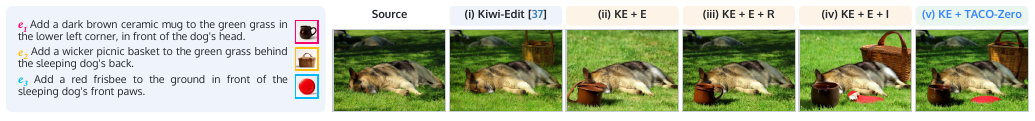}
  \vspace{-7.5mm}
  \caption{\textbf{Ablation study on Kiwi-Edit (KE)~\cite{kiwiedit}} on a record with $k = n = 3$. KE in~(i) adds only the basket. Per-instruction encoding in~(ii) adds only the mug and loses the shadows. Reference
  assignment in~(iii) keeps the shadows but adds only the mug, and instruction assignment in~(iv) adds all three objects but loses the shadows. Only
  \method in~(v) adds all three and preserves the untouched parts.}
  \label{fig:ablation}
\end{figure*}

\paragraph{Implementation details.}
We port \method to Kiwi-Edit~\cite{kiwiedit} and JoyAI-Video-Edit~\cite{joyai}. Kiwi-Edit runs the 30-block
Wan2.2-TI2V-5B DiT~\cite{wan} with Qwen2.5-VL-3B~\cite{qwen25vl} and renders
$832\times480$ videos in 50 steps, and we compute each $M_i$ at step $s=5$. JoyAI-Video-Edit is a 16B, 40-block
MMDiT with MiMo-VL-7B~\cite{mimovl}. It renders $840\times480$ videos one
latent frame at a time, in two distilled steps per frame, so we compute $M_i$
at $s=1$ in each latent frame. On both, the weights stay frozen, $F_i$ averages the top $L=10$
blocks, and we apply reference and instruction assignment in every DiT block at
every step. All methods edit 81 frames on one NVIDIA B200 GPU. More implementation details are in \supp{sec:app:impl}.

\subsection{Results}
\label{sec:exp:main}

\paragraph{Quantitative comparison.}
\label{sec:exp:quant}
\begingroup%
\Cref{tab:main} compares all methods on \bench. No method lands most records in full (shaded columns). The best baseline, CoinVE-Edit, lands $61.89\%$
of the instructions (\mt{LND}), but completes every instruction in only $27.20\%$
of the records (\mt{ALL}). Seedance~2.5 is next in \mt{LND} (57.93), Bernini in \mt{ALL} (25.40). Each baseline also has a weak point. CoinVE-Edit takes no reference, so
its \mt{RA} is low (59.26). LTX-Ripple matches references best (\mt{RA} 82.42)
but changes the rest of the video (\mt{SSO} 0.553). EditVid edits frame by
frame, so its motion is less smooth (\mt{MSM} 0.714). ReCo degrades with every
pass (\mt{SSO} 0.296). Kiwi-Edit-Ref keeps the source (\mt{SSO} 0.714) only
because it executes the fewest instructions (\mt{EXE} 50.60).

\method improves both backbones without training (blue vs.\ green rows). On Kiwi-Edit, it executes more instructions (\mt{EXE} 61.27
$\rightarrow$ 74.60) and binds them better (\mt{BND} 34.46 $\rightarrow$
59.77, \mt{RA} 32.27 $\rightarrow$ 47.26). \mt{LND} rises from 24.51 to
37.20, and \mt{ALL} from 5.40 to 10.00. \mt{SSO} dips (0.531
$\rightarrow$ 0.505), partly because Kiwi-Edit leaves edits undone: its
\mt{SSO} is 0.674 where it executes no instruction and 0.496 where it executes
all. JoyAI-Video-Edit already binds well (\mt{BND} 77.38), so there is less to fix: \mt{LND} rises from 54.11 to 57.37 and \mt{ALL} from 20.80 to 23.00.
Its \mt{SSO} rises (0.740 $\rightarrow$ 0.783), since each edit stays in
its own region. \mt{RS} stays at each backbone's level
(0.617 vs 0.619, 0.635 vs 0.638).%
\endgroup

\paragraph{Qualitative comparison.}
\begingroup%
\Cref{fig:qual} compares three records of growing $k$. Arrows mark some, not all, of the edits
that land (green), edits that are missed (red) and partial edits or unwanted
changes (orange). On the street (blue panel, $k = 3$), both backbones erase the
streetlight, and \method keeps it as the black pole the instruction asks for. At the
table (orange panel, $k = 4$), Kiwi-Edit changes the whole scene and adds no candelabra, while \method
preserves the scene and adds the candles. JoyAI-Video-Edit keeps the wooden bowl
and adds no candelabra, and \method lands all four edits. In the kitchen (green
panel, $k = 5$), both backbones keep the steel fridge, and \method turns it mint.
On Kiwi-Edit, \method also removes the second fridge.
The baselines show the trade-offs of \cref{tab:main}. LoomVideo removes the brick
building, in line with its low \mt{SSO} (0.582). LTX-Ripple paints the wall cabinets
mint and keeps the wooden counter (\mt{SSO} 0.553). Seedance~2.5 keeps the black pendant lamp in the kitchen. CoinVE-Edit adds no candelabra and merges both fridges.
Bernini turns part of the brick building into glass.
More qualitative results are in
\supp{sec:app:results}.%
\endgroup

\paragraph{Ablations.}
\label{sec:exp:ablation}

\begingroup%
We ablate \method on Kiwi-Edit (KE) over all 500 records (\cref{tab:main}) and
one video (\cref{fig:ablation}), and track two failures: edits that do not land
(\mt{LND}) and changes to the untouched parts of the video (\mt{SSO}). With the
joint prompt, KE lands few edits (\mt{LND} 24.51, \mt{SSO} 0.531) and adds only
the basket (\mbox{\cref{fig:ablation}-(i)}). \emph{Per-instruction encoding} (KE + E) worsens both. Each
region spreads its attention over all $k$ contexts (\cref{sec:analysis}), so
fewer edits land (\mt{LND} 19.94) and the untouched parts change (\mt{SSO}
0.460): only the mug is added, and the tree shadows are lost (\mbox{\cref{fig:ablation}-(ii)}).
\emph{Reference assignment} (KE + E + R) fixes the second failure: it confines
each reference to its own region, so references no longer leak into the rest of
the video. The untouched parts are kept (\mt{SSO} 0.532), shadows included
(\mbox{\cref{fig:ablation}-(iii)}), but few edits land (\mt{LND} 23.26). \emph{Instruction assignment} (KE + E + I) fixes
the first. Each region follows its own instruction, so more edits land
(\mt{LND} 33.60) and all three objects appear (\mbox{\cref{fig:ablation}-(iv)}), but the untouched parts
still change (\mt{SSO} 0.438). \emph{Our full method} (KE + \method) fixes both
failures: the most edits land (\mt{LND} 37.20, \mt{ALL} 5.40 $\rightarrow$
10.00), and the untouched parts stay close to KE's (\mt{SSO} 0.505), with all
three objects and their shadows (\mbox{\cref{fig:ablation}-(v)}).%
\endgroup

\begin{figure}[t]
  \centering
  \includegraphics[width=\columnwidth]{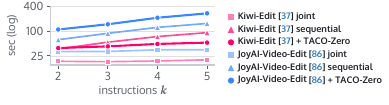}
  \vspace{\dimexpr -6mm-9pt\relax}
  \caption{\textbf{Runtime per record by $k$.} \method grows slowly with $k$ on Kiwi-Edit and faster on the few-step JoyAI-Video-Edit.}%
  \label{fig:runtime}
\end{figure}

\paragraph{Scaling with $k$.}
\begingroup%
We analyze the effect of increasing $k$ in \mbox{\cref{fig:teaser}-(D)}. On Kiwi-Edit, the
joint prompt lands 40.40\% of the edits at $k = 2$ but 18.88\% at $k = 5$,
and sequential editing falls from 15.60\% to 2.40\% as its artifacts
accumulate. \method lands the most edits at every $k$ and declines the least,
from 47.60\% to 34.88\%, since each instruction keeps its context. At
$k = 5$, it lands nearly twice as many as the joint prompt. On
JoyAI-Video-Edit, \method also leads at every $k$ (65.20\% to 55.52\%,
against 59.20\% to 52.00\%).%
\endgroup

\paragraph{Scaling with $n$.}
\begingroup%
We analyze the effect of increasing $n$ in \mbox{\cref{fig:teaser}-(D)}. On Kiwi-Edit, the
joint prompt lands 34.65\% of the edits at $n = 1$ but 7.20\% at $n = 5$, and
sequential editing falls from 8.51\% to 0\%. \method lands the most edits at
every $n$ and declines slightly (39.21\% to 36.00\%), since each reference stays with
its own instruction. At $n = 5$, it lands five times as many as the joint
prompt. On JoyAI-Video-Edit, \method
leads or ties at every $n$ (58.22\% to 63.20\%, against 53.07\% to 56.80\%).%
\endgroup

\paragraph{Runtime.}
\begingroup%
We report the runtime per record for $k = 2$ to 5 in
\cref{fig:runtime}. On Kiwi-Edit, \textit{Locate} stops each of its $k$ passes
after a tenth of the steps: \method takes 39 to 53 seconds, against 18 to 20 for
the joint prompt, and grows slower with $k$ than sequential editing (37 to 93),
while landing the most edits (\cref{tab:main}, \mbox{\cref{fig:teaser}-(D)}). On the few-step JoyAI-Video-Edit, a \textit{Locate} pass
is a full render, so \textit{Locate} adds more: 111 to 276 seconds, against 32 to 36
for the joint prompt and 63 to 156 for sequential editing. \method adds little memory. Full runtime and memory analysis is in \supp{sec:app:results}.%
\endgroup

\section{Conclusion}
\label{sec:conclusion}

We analyzed why a frozen video editor fails at compositional editing and proposed \method, to our knowledge, the first
zero-shot method that lets a single-instruction video editor compose multiple referenced edits. It encodes each instruction
separately, locates its target region, and assigns each region's tokens to its own instruction and reference.
We also built \bench, 500 records balanced over two to five instructions and one to five references. \method lifts landed edits on a cross-attention DiT and an MMDiT.

{
    \small
    \bibliographystyle{ieeenat_fullname}
    \bibliography{main}
}

\ifnum\showsupp=1
\clearpage
\global\suppstartedtrue
\makeatletter\ifnum\showmain=0 \global\let\protected@write\TZ@protectedwrite\fi\makeatother
\onecolumn%
\begin{center}\Large\textbf{\thetitle}\\[0.5em]Supplementary Material\\[1.0em]\end{center}%
\definecolor{figorange}{HTML}{FB5607}%

\appendix

\raggedbottom
\makeatletter
\newif\ifsupp@first
\newcommand\suppnobreak{\global\supp@firsttrue}
\newcommand\suppsec{\global\supp@firsttrue\supp@section}
\let\supp@section\section
\let\supp@subsection\subsection
\let\suppfloat\figure\let\endsuppfloat\endfigure
\let\supptab\table\let\endsupptab\endtable
\renewenvironment{table}[1][]{\supptab[!htp]}{\endsupptab}
\renewenvironment{table*}[1][]{\supptab[!htp]}{\endsupptab}
\renewenvironment{figure}[1][]{\par\medskip\noindent\begin{minipage}{\linewidth}\def\@captype{figure}}{\end{minipage}\par\medskip}
\renewenvironment{figure*}[1][]{\par\medskip\noindent\begin{minipage}{\linewidth}\def\@captype{figure}}{\end{minipage}\par\medskip}
\makeatother
\renewcommand\topfraction{0.95}\renewcommand\textfraction{0.05}\renewcommand\floatpagefraction{0.75}\setcounter{topnumber}{3}\setcounter{totalnumber}{5}
\interlinepenalty=10000 \clubpenalty=10000 \widowpenalty=10000 \displaywidowpenalty=10000
\postdisplaypenalty=10000 \brokenpenalty=10000

\FloatBarrier
\makeatletter
\newcommand\supptocsec[2]{\par\vspace{2pt}\noindent\hyperref[#1]{\textbf{\ref*{#1}.\ \ #2}}\hfill\hyperref[#1]{\textbf{\pageref*{#1}}}\par}
\newcommand\supptocsub[2]{\par\noindent\hspace{1.2em}\hyperref[#1]{\ref*{#1}\ \ #2}\ \dotfill\ \hyperref[#1]{\pageref*{#1}}\par}
\makeatother
\begingroup\hypersetup{linkcolor=black}%
\begin{center}
{\large\textbf{Contents}}\par\vspace{16pt}
{\small
\begin{minipage}{0.75\linewidth}
  \supptocsec{sec:app:analysis}{Analysis details}
  \supptocsub{sec:app:analysis:record}{The record of Fig.~2}
  \supptocsub{sec:app:analysis:all}{Over all records}
  \supptocsec{sec:app:impl}{More implementation details}
  \supptocsub{sec:app:mask}{Mask details}
  \supptocsub{sec:app:hosts}{Backbones}
  \supptocsub{sec:app:hparams}{Hyperparameters}
  \supptocsub{sec:app:baselines}{Baseline inputs}
  \supptocsec{sec:app:bench}{\bench details}
  \supptocsub{sec:app:gap}{Existing benchmarks}
  \supptocsub{sec:app:clips}{Clip screening and selection}
  \supptocsub{sec:app:refs}{Instructions and references}
  \supptocsub{sec:app:checklist}{Checklist questions}
  \supptocsub{sec:app:human}{User study details}
  \supptocsec{sec:app:results}{Additional results}
  \supptocsub{sec:app:cells}{\mt{LND} for every pair of $k$ and $n$}
  \supptocsub{sec:app:order}{Robustness to instruction order}
  \supptocsub{sec:app:sam}{Attention masks against SAM~3.1 masks}
  \supptocsub{sec:app:rsbound}{Upper bound for referenced instructions}
  \supptocsub{sec:app:xroute}{More ablations}
  \supptocsub{sec:app:textonly}{\method on another public benchmark}
  \supptocsub{sec:app:cost}{Runtime and memory}
  \supptocsub{sec:app:qual}{More qualitative results}
  \supptocsec{sec:app:limits}{Limitations}
\end{minipage}\par}
\end{center}
\endgroup
\clearpage

\section{Analysis details}
\label{sec:app:analysis}

\subsection{The record of \texorpdfstring{\cref{fig:analysis}}{Fig. 2}}
\label{sec:app:analysis:record}
This section gives the details of \cref{fig:analysis}. The record has three instructions: add a bench on the grass,
remove the back horse, and swap the chestnut horse for the pony of the reference. Every panel uses the same seed. Panels
(B) and (C) show the cross-attention at step 5, averaged over blocks 13 to 16 and over all heads. The region of $e_i$ is
the set $M_i$ of video tokens that \textit{Locate} gives $e_i$ (\cref{sec:ours:locate}). For panel (B), we take the 128
tokens of the joint prompt that each region attends to most, and we count how many of these tokens two regions share. The
three pairs of regions share 117, 120 and 120 of the 128 tokens, so all three regions attend to almost the same tokens.
For panel (C), each bar splits the attention of one region over the contexts $c_1$, $c_2$ and $c_3$, and the outlined part
of the bar is the share that goes to the region's own context. The context $c_3$ of the pony swap holds the reference, so
$c_3$ has 1,497 tokens, against 512 each for $c_1$ and $c_2$. This record is not one of the 500 records of \bench.
\Cref{sec:app:analysis:all} measures the same quantities on all 500 records.

\subsection{Over all records}
\label{sec:app:analysis:all}
We repeat the analysis of \cref{sec:analysis} on \bench, with Kiwi-Edit~\cite{kiwiedit}, the same step, blocks and heads, and
the regions $M_i$ of \method. A whole-frame instruction has no region and is left out. (B) and (C) need two regions, which
490 of the 500 records have. \Cref{tab:app:analysis,tab:app:analysis:b,tab:app:analysis:c} report the three findings by $k$.

\paragraph{(A) A single instruction finds its target.} \Cref{tab:app:analysis} compares the region $M_i$ of each instruction
with the SAM~3.1 mask of the instruction's target, over the 834 instructions whose target SAM~3.1 finds on the source video.
We report two measures: the IoU of $M_i$ with the target mask, and the coverage, which is the share of the target inside
$M_i$. Both measures grow with the size of $M_i$, so we compare each with a chance level: we shift the same $M_i$ to a random
place in the frame, 20 times, and measure again. The median IoU is 0.22, against 0.04 at a random place, and 93\% of the
instructions are above their own chance level. The median coverage is 52\%, against 10\% at a random place. A region thus
holds about half of its target, and the IoU stays low because the region also covers some of the surroundings. The edit
also lands: alone, Kiwi-Edit lands 56.95\% of the instructions with a reference, against 21.30\% with the joint prompt
(\cref{tab:app:rsbound}).

\begin{table}[!ht]
  \centering
  \caption{\textbf{(A) A single instruction finds its target} (Kiwi-Edit, step 5, medians over the 834 instructions whose
  target SAM~3.1 finds on the source video). IoU: overlap of each region $M_i$ with the SAM~3.1 mask of its target. Coverage:
  the share of the target inside $M_i$. Random: the same $M_i$ shifted to a random place in the frame, averaged over 20
  shifts. All masks are on the latent grid.}
  \label{tab:app:analysis}
  \supptabfont\setlength{\tabcolsep}{6pt}
  \begin{tabular}{@{}ccc*{4}{c}@{}}
    \toprule
    & & & \multicolumn{2}{c}{IoU} & \multicolumn{2}{c}{Coverage} \\
    \cmidrule(lr){4-5}\cmidrule(l){6-7}
    $k$ & Records & Instructions & $M_i$ & Random & $M_i$ & Random \\
    \midrule
    2 & 97 & 131 & 0.19 & 0.06 & 0.39 & 0.11 \\
    3 & 109 & 169 & 0.21 & 0.05 & 0.54 & 0.11 \\
    4 & 116 & 228 & 0.22 & 0.04 & 0.56 & 0.10 \\
    5 & 119 & 306 & 0.23 & 0.03 & 0.55 & 0.09 \\
    \bottomrule
  \end{tabular}
\end{table}

\paragraph{(B) The joint prompt entangles the instructions.} With the joint prompt, all instructions share one context.
\Cref{tab:app:analysis:b} tests in two ways whether a region reads only its own instruction. First, we take the 128 context
tokens that each region attends to most. Two regions share a median of 118 to 120 of these 128 tokens, so the record of
\cref{fig:analysis} is typical (114 to 119 with all references in the prompt, as here).
Second, we replace one instruction $e_j$ in the joint prompt with a neutral instruction, and we measure how much the
cross-attention output changes, relative to its norm. The output changes by 0.39 to 0.72 in the region of $e_j$ and by 0.36
to 0.57 in the regions of the other instructions, so the other regions change almost as much as the region of $e_j$.
What each region receives therefore depends on every instruction, not only its own.

\begin{table}[!ht]
  \centering
  \caption{\textbf{(B) The joint prompt entangles the instructions} (Kiwi-Edit, joint prompt with all references, step 5,
  medians). Shared of 128: context tokens that two regions share among the 128 each attends to most. Change: the
  cross-attention output change, relative to its norm, when one instruction $e_j$ is replaced with a neutral one, in
  the region of $e_j$ (Own) and in the regions of the other instructions (Other).}
  \label{tab:app:analysis:b}
  \supptabfont\setlength{\tabcolsep}{6pt}
  \begin{tabular}{@{}cc*{3}{c}@{}}
    \toprule
    & & Shared of 128 & \multicolumn{2}{c}{Change} \\
    \cmidrule(l){4-5}
    $k$ & Records & Regions & Own & Other \\
    \midrule
    2 & 115 & 118 & 0.72 & 0.57 \\
    3 & 125 & 119 & 0.56 & 0.45 \\
    4 & 125 & 119 & 0.45 & 0.39 \\
    5 & 125 & 120 & 0.39 & 0.36 \\
    \bottomrule
  \end{tabular}
\end{table}

\paragraph{(C) Concatenated contexts split the attention.} With concatenated contexts, each instruction $e_i$ has its own
context $c_i$. \Cref{tab:app:analysis:c} measures how the video tokens of a region divide their cross-attention among the $k$
contexts. If a region followed its own instruction, it would give its own context most of its attention. It does not: a
region gives its own context a median share of 0.503, 0.336, 0.251 and 0.201 for $k = 2$ to 5, which is the even split
$1/k$. The other regions give the same context almost the same share, so a region prefers its own context by only 0.1 to
0.6 percentage points. The contexts are separate, but nothing tells a video token which context to follow. This is why
\method needs instruction assignment (\cref{sec:ours:render}).

\begin{table}[!ht]
  \centering
  \caption{\textbf{(C) Concatenated contexts split the attention} (Kiwi-Edit, step 5, medians over records). The share of a
  region's attention on its own context $c_i$ (Own), and the share that the other regions give $c_i$ (Others), against $1/k$.}
  \label{tab:app:analysis:c}
  \supptabfont\setlength{\tabcolsep}{6pt}
  \begin{tabular}{@{}ccccc@{}}
    \toprule
    $k$ & Records & Own & Others & $1/k$ \\
    \midrule
    2 & 115 & 0.503 & 0.497 & 0.500 \\
    3 & 125 & 0.336 & 0.333 & 0.333 \\
    4 & 125 & 0.251 & 0.250 & 0.250 \\
    5 & 125 & 0.201 & 0.200 & 0.200 \\
    \bottomrule
  \end{tabular}
\end{table}

\section{More implementation details}
\label{sec:app:impl}

\subsection{Mask details}
\label{sec:app:mask}
We complete \textit{Locate} (\cref{sec:ours:locate}) with three details.

\paragraph{Smoothing.} The Gaussian $G$ of \cref{eq:contrast} has $\sigma = 0.5$ latent frames in time and
$\sigma = 1.0$ latent cells in space.

\paragraph{Cleanup.} After thresholding, a candidate set $\tilde{M}_i$ can contain a few stray tokens far from the target
of $e_i$. We remove every group of fewer than 4 connected tokens from $\tilde{M}_i$. This cleanup runs before the overlap
rule of \cref{eq:mask}.

\paragraph{Whole-frame instructions.} A style instruction edits the whole frame, so it has no region to locate. \bench has
such an instruction in 94 of its 500 records. Each one says ``to the whole frame'', and we find them by their edit type.
A whole-frame instruction gets no mask: \cref{eq:mask} builds masks for the other instructions only, and \cref{eq:xroute}
never blocks the context of a whole-frame instruction. As a result, a video token in $M_j$ attends to its own context
$c_j$ and to every whole-frame context. A video token in no mask attends to every context.

\subsection{Backbones}
\label{sec:app:hosts}
Kiwi-Edit and JoyAI-Video-Edit hold instructions and references differently, so we apply \method to each as follows.
Both backbones encode each $e_i$ with the message of one sequential pass (\cref{fig:app:inputs}).

\paragraph{Kiwi-Edit.} Kiwi-Edit pads each reference to the video's shape, which is why \cref{eq:route} blocks padding
tokens.

\paragraph{JoyAI-Video-Edit.} JoyAI-Video-Edit has no padding. It keeps each reference in its attention cache, so
\cref{eq:route} blocks the cache tokens of the other references. Its MMDiT computes one joint attention over video and
context tokens, so both assignments act inside this joint attention, and only on the noisy video tokens. The source
tokens, which JoyAI-Video-Edit concatenates to the noisy tokens (\cref{eq:input}), keep every context and every reference.
In the joint attention, the context tokens also attend to each other. We restrict each context to itself, which keeps
the $k$ contexts from mixing.

\subsection{Hyperparameters}
\label{sec:app:hparams}
\Cref{tab:app:hparams} lists the settings of \method on each backbone.

\paragraph{Masks.} Kiwi-Edit denoises in 50 steps, and we read the masks after $s = 5$ steps, a tenth of the steps.
JoyAI-Video-Edit renders one latent frame at a time in two distilled steps, so we read the masks at the first step of
each latent frame. On both backbones, the attention maps average the $L = 10$ blocks with the highest $\delta_l$.

\paragraph{Assignments.} On both backbones, reference assignment and instruction assignment act in every block and at
every step. The numbers of steps with reference assignment and with instruction assignment, $\eta_R$ and $\eta_I$,
therefore equal the number of denoising steps. \Cref{sec:app:xroute} varies $s$, $L$, $\eta_R$ and $\eta_I$.

\begin{table}[!ht]
  \centering
  \caption{\textbf{Hyperparameters of \method on each backbone.} The weights of both backbones stay frozen.}
  \label{tab:app:hparams}
  \supptabfont\setlength{\tabcolsep}{5pt}
  \begin{tabular}[t]{@{}lcc@{}}
    \toprule
    Backbone & Kiwi-Edit~\cite{kiwiedit} & JoyAI-Video-Edit~\cite{joyai} \\
    \midrule
    DiT & Wan2.2-TI2V-5B~\cite{wan} & MMDiT~\cite{sd3}, 16B \\
    DiT blocks & 30 & 40 \\
    MLLM & Qwen2.5-VL-3B~\cite{qwen25vl} & MiMo-VL-7B~\cite{mimovl} \\
    Resolution & $832\times480$ & $840\times480$ \\
    Frames & 81 & 81 \\
    Denoising steps & 50 & 2 per latent frame \\
    Seed & 0 & 0 \\
    \bottomrule
  \end{tabular}\hspace{18pt}%
  \begin{tabular}[t]{@{}lcc@{}}
    \toprule
    \method & Kiwi-Edit~\cite{kiwiedit} & JoyAI-Video-Edit~\cite{joyai} \\
    \midrule
    Mask step $s$ & 5 & 1 per latent frame \\
    Mask blocks $L$ & 10 & 10 \\
    Gaussian $G$, $\sigma$ (time, space) & 0.5 frames, 1.0 cells & 0.5 frames, 1.0 cells \\
    Threshold $\gamma_i$ & Otsu~\cite{otsu} & Otsu~\cite{otsu} \\
    Smallest token group kept & 4 & 4 \\
    Reference assignment blocks & all 30 & all 40 \\
    Reference assignment steps $\eta_R$ & 50 & 2 per latent frame \\
    Instruction assignment blocks & all 30 & all 40 \\
    Instruction assignment steps $\eta_I$ & 50 & 2 per latent frame \\
    \bottomrule
  \end{tabular}
\end{table}

\subsection{Baseline inputs}
\label{sec:app:baselines}
\Cref{fig:app:inputs} shows the input of every editor. Most baselines take one instruction, so we adapt the input of each
baseline to $k$ instructions and $n$ references.

\paragraph{Joint prompt.} The \emph{joint prompt} numbers the $k$ instructions and ends with one sentence per reference,
``Reference image $j$ is for instruction $i$''. Bernini~\cite{bernini} (\mbox{\cref{fig:app:inputs}-(A)}) and EditVid~\cite{editvid} (\mbox{\cref{fig:app:inputs}-(B)}) take the joint
prompt and all $n$ references unchanged. The stock MLLM message of Kiwi-Edit~\cite{kiwiedit} (\mbox{\cref{fig:app:inputs}-(C)}) and
Kiwi-Edit-Ref~\cite{kiwiedit} (\mbox{\cref{fig:app:inputs}-(D)}) holds one reference, so we extend the message to all $n$ references, each after the label
``Reference image $j$:''. The MLLM of JoyAI-Video-Edit~\cite{joyai} (\mbox{\cref{fig:app:inputs}-(E)}) never reads a reference, so we encode each
reference alone and join the encoded references in the attention cache of JoyAI-Video-Edit.

\paragraph{Other input formats.} Seedance~2.5~\cite{seedance25} (\mbox{\cref{fig:app:inputs}-(F)}) and LoomVideo~\cite{loomvideo} (\mbox{\cref{fig:app:inputs}-(G)}) point to references
in their own syntax: ``(@Image $j$)'' follows the object that reference $j$ shows. MiniMax-H3~\cite{minimaxh3} (\mbox{\cref{fig:app:inputs}-(H)}) expects
a six-section prompt, so we fill the vendor's template with the numbered instructions. LTX-Ripple~\cite{ltxripple} (\mbox{\cref{fig:app:inputs}-(I)}) has
two stages: FLUX.2-Klein~\cite{flux2} edits the first frame from the joint prompt and all references, and LTX-Ripple
propagates the edited frame without a reference. CoinVE-Edit~\cite{coinve} (\mbox{\cref{fig:app:inputs}-(J)}) encodes each instruction separately and
takes no reference.

\paragraph{Sequential editing.} \emph{Sequential} editing runs the editor $k$ times. Each pass edits the output of the
previous pass with one instruction and the reference of that instruction. We run sequential editing on Kiwi-Edit (\mbox{\cref{fig:app:inputs}-(K)}),
JoyAI-Video-Edit (\mbox{\cref{fig:app:inputs}-(L)}) and ReCo~\cite{reco} (\mbox{\cref{fig:app:inputs}-(M)}).

\begin{suppfloat}[tp]
  \centering
  \includegraphics[width=\textwidth]{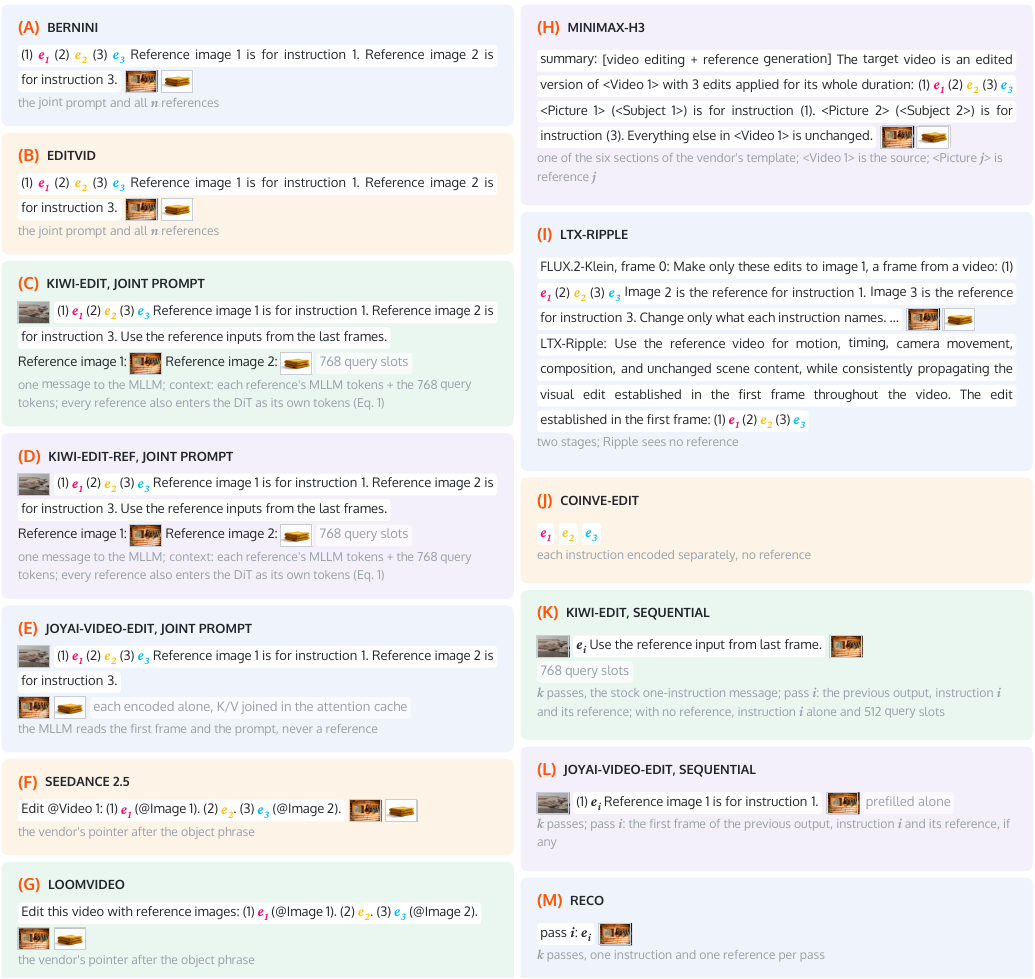}
  \vspace{\dimexpr -3mm-8.3pt\relax}%
  \caption{\textbf{Baseline input formats}, for $k = 3$ instructions $e_1, e_2, e_3$ and $n = 2$ references, reference 1 for
  $e_1$ and reference 2 for $e_3$. Thumbnails show the source video (the first frame for JoyAI-Video-Edit) and the references; dark text is
  the prompt, and grey text describes the rest of the input. (A)~Bernini~\cite{bernini} and (B)~EditVid~\cite{editvid} take the joint prompt with all references.
  (C)~Kiwi-Edit~\cite{kiwiedit} and (D)~Kiwi-Edit-Ref~\cite{kiwiedit} take the joint prompt with each reference after its label, and (E)~JoyAI-Video-Edit~\cite{joyai}
  takes it with each reference encoded alone. (F)~Seedance~2.5~\cite{seedance25} and (G)~LoomVideo~\cite{loomvideo} use their own ``(@Image $j$)'' pointers, and
  (H)~MiniMax-H3~\cite{minimaxh3} a six-section template, of which we show the summary. (I)~LTX-Ripple~\cite{ltxripple} has one prompt for each of its two stages.
  (J)~CoinVE-Edit~\cite{coinve} takes each instruction alone, with no reference. (K)~Kiwi-Edit, (L)~JoyAI-Video-Edit and (M)~ReCo~\cite{reco} run sequentially,
  where $e_i$ is the instruction of pass $i$; ReCo takes one instruction and one reference per pass.}
  \label{fig:app:inputs}
\end{suppfloat}

\makeatletter
\renewcommand\section{\global\supp@firsttrue\supp@section}
\renewcommand\subsection{\global\supp@firstfalse\supp@subsection}
\makeatother
\section{\bench details}%
\label{sec:app:bench}

\subsection{Existing benchmarks}
\label{sec:app:gap}
\Cref{sec:bench} names two needs that a benchmark for our setting must meet: (i) it varies $k$ and $n$, and (ii) it tells
which instruction failed. This section checks existing benchmarks against both, and \cref{tab:bench-landscape} sums them up.
Unless a paper states it, an entry is our reading of the released records.

\paragraph{Reference benchmarks.} RefVIE-Bench~\cite{kiwiedit} has references but one instruction per record.
OmniEdit-Bench~\cite{omnieditbench} has 200 reference cases of 790, one instruction each.
Goku-Bench~\cite{goku} keeps reference edits and multi-step edits in disjoint subsets.

\paragraph{Compositional benchmarks.} CoVEBench~\cite{covebench} and CoinVE-Bench~\cite{coinve} give several instructions per
video but no references, and their records thin out as $k$ grows. The 626 records of CoVEBench carry 1 to 6 edits (43, 195,
206, 152, 28 and 2 records, respectively). CoinVE-Bench has 143, 108, 74 and 36 records at $k = 2$ to 5 and no image field.
The two differ in how they score. CoVEBench groups its questions by edit and scores each group on its own, so it tells which
edit failed. In CoinVE-Bench, the field that ties a question to an instruction is unset on $2{,}470$ of $2{,}602$ question
groups, so most of its questions cannot be traced to one instruction.

\paragraph{Multi-reference benchmarks.} UniVBench~\cite{univbench} pairs 200 videos with instructions and reference
images, and its release has since grown to 620 videos. A record has one compound instruction that names its references by number, and the records drop from
573 with two references to 44 with three. TIDE-Bench~\cite{tide} has 211 scenarios, each with one instruction and up to three
references, and it records which element each reference shows. Both write all edits and references into one instruction, and neither controls $k$ or $n$.
Neither scores each instruction on its own: UniV-Eval rates each shot on eight dimensions, and TIDE-Bench gives one Edit
Completeness score to all edits of a scenario and keeps the lowest of its per-reference scores, so a failed edit cannot be
traced to its instruction. ReBind~\cite{rebind} also binds its
references inside one dense instruction.

\paragraph{Summary.} No existing benchmark meets both needs: the one that scores each edit on its own, CoVEBench, has no
references, and those with several references per record control neither $k$ nor $n$. \bench holds 125 records for each $k$ from 2 to 5, spread over
$n = 1$ to $k$, binds each reference to its own instruction, and asks questions per instruction (\cref{sec:bench:scoring}).

\begin{table}[!ht]
  \centering
  \caption{\textbf{Comparison with existing benchmarks.} \emph{Multiple instr.}: several instructions per record. \emph{Ref.\ per
  instr.}: each reference belongs to one instruction. \emph{$k$, $n$ balanced}: records spread evenly over $k$ and $n$.
  \emph{Per-instr.\ score}: each instruction is scored on its own; \emph{partial}: most questions of CoinVE-Bench cannot be
  traced to one instruction (\cref{sec:app:gap}). n/a: no record has both several instructions and a reference.
  \emph{\#~Videos}: source videos. \emph{\#~Records}: evaluation records; for UniVBench, its multi-reference editing set.
  $^\dagger$References and multi-step edits are in disjoint subsets. $^\ddagger$Counted on the released files; the papers
  report 110 records for RefVIE-Bench and 200 videos for UniVBench, and give no count for the other marked entries.
  $^\S$Not released; from the paper. The marks are our reading of the papers and released records.}
  \label{tab:bench-landscape}
  \supptabfont\setlength{\tabcolsep}{6.5pt}%
  \begin{tabular}{lcccccccc}
    \toprule
    Benchmark & Video & Multiple instr. & References & Ref.\ per instr. & $k$, $n$ balanced & Per-instr.\ score & \# Videos & \# Records \\
    \midrule
    \multicolumn{9}{l}{\textbf{\textit{Reference}}} \\
    RefVIE-Bench~\cite{kiwiedit}          & \cmark & \xmark & \cmark & n/a    & \xmark & n/a     & 32$^\ddagger$ & 120$^\ddagger$\\
    OmniEdit-Bench~\cite{omnieditbench}   & \cmark & \xmark & \cmark & n/a    & \xmark & n/a     & 790$^\ddagger$ & 790\\
    Goku-Bench~\cite{goku}                & \cmark & \cmark & \cmark$^\dagger$ & n/a$^\dagger$ & \xmark & \xmark & 1{,}000$^\S$ & 1{,}000$^\S$\\
    \midrule
    \multicolumn{9}{l}{\textbf{\textit{Compositional}}} \\
    CoVEBench~\cite{covebench}            & \cmark & \cmark & \xmark & \xmark & \xmark & \cmark  & 416 & 626\\
    CoinVE-Bench~\cite{coinve}            & \cmark & \cmark & \xmark & \xmark & \xmark & partial & 361 & 361$^\ddagger$\\
    \midrule
    \multicolumn{9}{l}{\textbf{\textit{Multi-reference}}} \\
    UniVBench~\cite{univbench}            & \cmark & \cmark & \cmark & \xmark & \xmark & \xmark  & 620$^\ddagger$ & 620$^\ddagger$\\
    TIDE-Bench~\cite{tide}                & \cmark & \cmark & \cmark & \cmark$^\S$ & \xmark & \xmark & 56$^\S$ & 211$^\S$\\
    \midrule
    \rowcolor{oursrow}
    \bench                                & \cmark & \cmark & \cmark & \cmark & \cmark & \cmark  & 330 & 500\\
    \bottomrule
  \end{tabular}
\end{table}

\suppnobreak
\subsection{Clip screening and selection}
\label{sec:app:clips}
This section details how we screen and select the 330 clips of \cref{sec:bench:construction}. We screen $1{,}640$
Pexels~\cite{pexels} clips in eight stages, S1 to S8 (\mbox{\cref{fig:app:stats}-(A)}).

\paragraph{Screening (S1 to S7).} S1 checks the format, 1080p and at least ten seconds, which every clip passes. S2 drops
the clips with a shot cut. S3 keeps a clip if SAM~3~\cite{sam3} finds at least four disjoint objects in the clip. S4 tracks
the objects of every clip through the best $5.06\,$s window of the clip and keeps the clip if at least two objects stay
visible. S4 runs on every clip, so it recovers clips that S3 rejected. S5 drops the clips with a watermark, such as a logo
or caption, burned into the image. S6 keeps the 345 clips that score at least 4 of 10 for relevance to our edit types and
pass a quality bar (DOVER++ technical 10~\cite{dover}, aesthetic 4.5~\cite{aesv25}). S7 labels the camera motion of each
clip and drops no clip.

\paragraph{Selection (S8).} From the 345 clips, a solver picks the 500 records that maximize relevance under the record
counts of \mbox{\cref{fig:app:stats}-(B)}: 125 records per $k$, split evenly over $n$. The solver allows at most two records per clip and two clips per
shoot. It also keeps floors on diversity (\mbox{\cref{fig:app:stats}-(G)}) and on each edit type (\mbox{\cref{fig:app:stats}-(D)}), so each edit type has $4.6\%$ to $28.9\%$
of the $1{,}750$ instructions. The solver floors quality but never ranks clips on quality, since we also report quality
as a metric (\mbox{\cref{fig:app:stats}-(C)}). A clip with two records gets a second instruction pool. We write the second pool with the
instructions of the first pool marked as used, so the two records share no instruction or reference. We check every
record by hand and keep or drop it.

\begin{suppfloat}[tp]
  \centering
  \newcommand{\statX}{1.0}%
  \newlength{\statg}\setlength{\statg}{0.045in}%
  \newlength{\statw}\setlength{\statw}{\dimexpr(\statX\textwidth-2\statg)/3-0.1pt\relax}%
  \newlength{\statv}\setlength{\statv}{\dimexpr(\statX\textwidth-\statg)/2-0.1pt\relax}%
  \begin{minipage}[b]{\statw}\includegraphics[width=\statw]{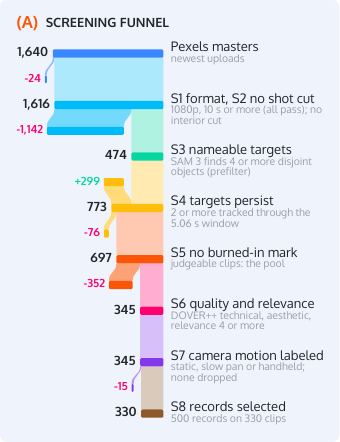}\end{minipage}\hspace{\statg}%
  \begin{minipage}[b]{\dimexpr 2\statw+\statg\relax}%
    \includegraphics[width=\statw]{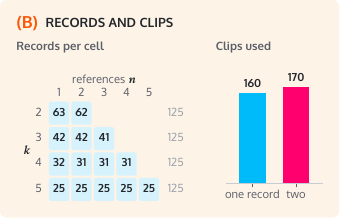}\hspace{\statg}\includegraphics[width=\statw]{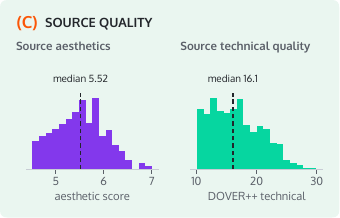}\\[\dimexpr\statg-1pt\relax]
    \includegraphics[width=\statw]{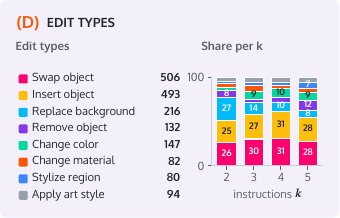}\hspace{\statg}\includegraphics[width=\statw]{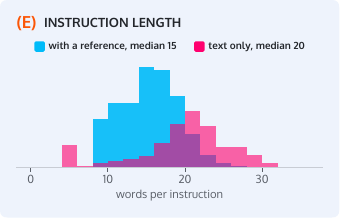}%
  \end{minipage}\\[\dimexpr\statg-1pt\relax]
  \includegraphics[width=\statw]{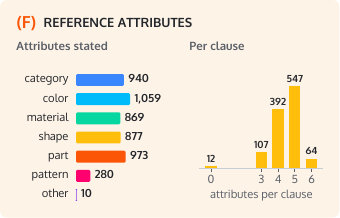}\hspace{\statg}\includegraphics[width=\statw]{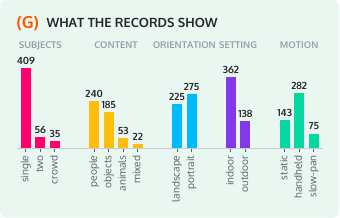}\hspace{\statg}\includegraphics[width=\statw]{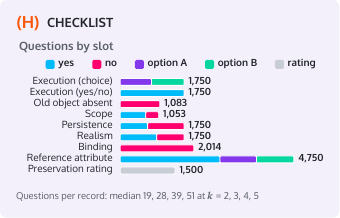}\\[\dimexpr\statg-1pt\relax]
  \includegraphics[width=\dimexpr 3\statw+2\statg\relax]{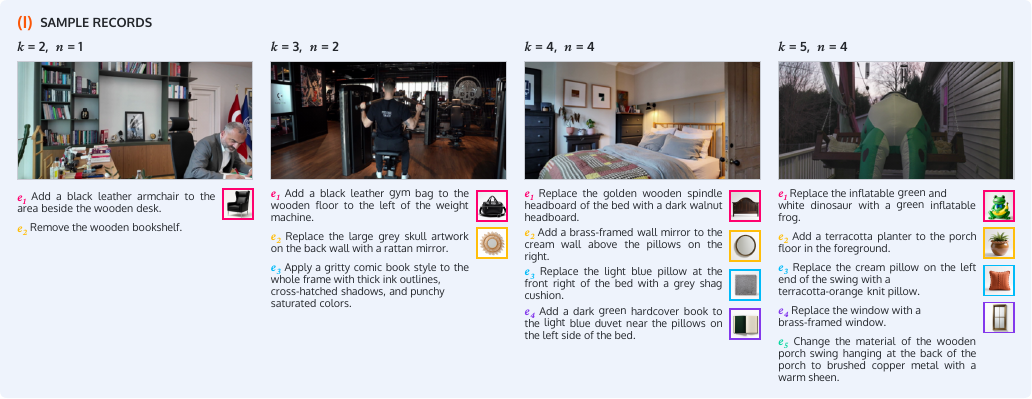}
  \vspace{\dimexpr -2mm-10.2pt\relax}
  \caption{\textbf{\bench statistics and sample records.} (A)~Clips kept at each screening stage, from the top down; bars branching to the left are the clips a stage drops. S4 tracks every clip, so it recovers clips S3 rejected. S8 selects 330 clips, which give the 500 records. (B)~125 records per $k$, split evenly over $n$; 170 of the 330 clips carry a second record. (C)~Source quality,
  floored at selection and never ranked on. (D)~Edit types of the $1{,}750$ instructions, overall and within each $k$.
  (E)~Instruction length with and without a reference. (F)~The attributes a reference description states; each becomes one
  \mt{RA} question. (G)~What the records show, as labeled at screening. (H)~Questions per record by $k$ and by slot and
  expected answer. (I)~One sample record for each $k$ from 2 to 5: the first frame and the instructions verbatim. More
  records are on our \webpage.}
  \label{fig:app:stats}
\end{suppfloat}

\suppnobreak
\subsection{Instructions and references}
\label{sec:app:refs}
\label{sec:app:instr}
This section details how we write and check the instructions and references of \cref{sec:bench:construction}.

\paragraph{Instructions.} Qwen3-VL-8B~\cite{qwen3vl} writes each instruction for one of the objects that
SAM~3~\cite{sam3} found. Each object gets at most one instruction, so no two instructions of a record edit
the same object. For each instruction, Qwen3-VL-8B first writes a detailed description, which states at least three
visual attributes when the instruction has a reference (\mbox{\cref{fig:app:stats}-(F)}). Qwen3-VL-8B then writes the
instruction that the editor reads, which starts with a verb and asks for one edit
(\mbox{\cref{fig:app:stats}-(D), (E)}). An instruction with a reference names its object only by class, color and
material.

\paragraph{References.} Qwen-Image~\cite{qwenimage} generates each reference from the detailed description, and
Qwen3-VL-8B checks that every stated attribute is visible in the reference. Only insertion, swap and background
replacement take a reference, and references go to the first instructions of a pool, so the mix of edit types changes
with $n$. At $k = n = 5$, for example, $54\%$ of the instructions are insertions, $42\%$ swaps and $3\%$ background
replacements.

\suppnobreak
\subsection{Checklist questions}
\label{sec:app:checklist}
This section gives the details of the checklist questions of \cref{sec:bench:scoring}: how many questions of each type
there are, which answer each type expects, and how we checked the questions.

\paragraph{Question counts.} The checklist holds $17{,}400$ questions, 14 to 72 per record, with medians of 19, 28, 39 and 51
at $k = 2$, 3, 4 and 5, respectively (\mbox{\cref{fig:app:stats}-(H)}). Each instruction gets four questions, $1{,}750$ of each
kind: two execution questions (one A/B, one yes/no), one persistence question and one realism question. An instruction that
removes or replaces something adds an absence question ($1{,}083$). A scope question asks about an untouched neighbor
($1{,}053$, since a neighbor shared by several instructions is asked once). Binding questions ask whether an edit leaked onto
the target of another edit ($2{,}014$), and each attribute of a reference adds one question ($4{,}750$ over the $1{,}122$
references). Three preservation ratings per record add $1{,}500$ and give \mt{SEM}.

\paragraph{Expected answers.} Persistence, realism and scope questions alternate their expected answer, execution and
reference questions always expect \emph{yes}, and binding and absence questions always expect \emph{no}. Overall, $57\%$ of
our yes/no questions expect \emph{yes}, against $74\%$ for CoVEBench~\cite{covebench} and $95\%$ for
CoinVE-Bench~\cite{coinve}. A judge that always answers \emph{no} would pass the binding and absence questions. We therefore
score a binding question zero when its execution question fails, and for an absence question the judge compares the result
with the source (\cref{sec:bench:scoring}).

\paragraph{Quality check.} We checked every generated question by hand and readjusted the template phrasing where a question
read unnaturally or asked about an edit that could not happen. Sample checklist questions are on our \webpage.

\subsection{User study details}
\label{sec:app:human}
This section checks the judge's answers to the checklist questions against people: the user study behind
\cref{sec:bench:scoring}, how often the judge and people give the same answer, and the scores that each gives every method.

\paragraph{Setup.} We draw 30 records spread over $k$ and $n$ and edit each with six methods: Seedance~2.5~\cite{seedance25},
CoinVE-Edit~\cite{coinve}, LTX-Ripple~\cite{ltxripple}, Kiwi-Edit~\cite{kiwiedit}, Kiwi-Edit with \method and
JoyAI-Video-Edit~\cite{joyai} with \method. This gives 180 edited videos. Each of 27 people rates 20 of the videos, so three
different people rate each video and no one sees a record twice. For each video, a person answers eight of the judge's
checklist questions for that record, spread over the question types, with \emph{yes} or \emph{no} or by picking one of two
options. A third button, \emph{the edit isn't there}, is for questions about an edit that was not made. The study runs in the
browser, on the page of \cref{fig:app:survey}: the source as Video~A and the edit as Video~B loop side by side in sync, with
the reference image when a question is about it. The page never names a method and shows no scores. Before the study, each
person reads a two-minute guide with an example clip of each artifact word, such as flicker and ghosting, and answers
practice questions with feedback. Each person also rates one check video, the unedited clip, whose answers are not counted.

\paragraph{Agreement with the judge.} As in CoVEBench~\cite{covebench}, we compare the raw answers of a person and the judge
by their exact-match rate and Cohen's $\kappa$~\cite{landis1977}, and we leave out the answers \emph{the edit isn't there}.
The judge gives the same answer as a single rater on 81.5\% of the questions ($\kappa = 0.63$). Where all three raters
answered a question, the judge gives the same answer as their majority on 85.2\% ($\kappa = 0.70$).

\paragraph{Human scores.} Every checklist question has an expected answer, the answer that a successful edit receives
(\cref{sec:app:checklist}). For each method, \cref{tab:app:human} reports the percentage of people's answers that are the
expected answer, over the 240 study questions on the 30 videos of that method, and the same percentage for the judge. Here
\emph{the edit isn't there} counts as a failed question, and, as in \cref{sec:bench:scoring}, the persistence, realism,
binding and reference questions of an instruction score zero when a person fails its execution question. People and the judge rank the six methods in the same order (Spearman's $\rho$ over six methods: 0.99 over all questions, 0.83 on the
execution and 0.84 on the realism questions), except that the judge ties CoinVE-Edit and LTX-Ripple. Adding \method to
Kiwi-Edit raises the human score from 52.5\% to 59.3\% and the judge's score from 54.2\% to 70.0\%. People give every method
a lower score than the judge does, by 4.1 points on average on the execution questions and by 14.2 on the realism questions.

\begin{suppfloat}[tp]
  \centering
  \includegraphics[width=\textwidth]{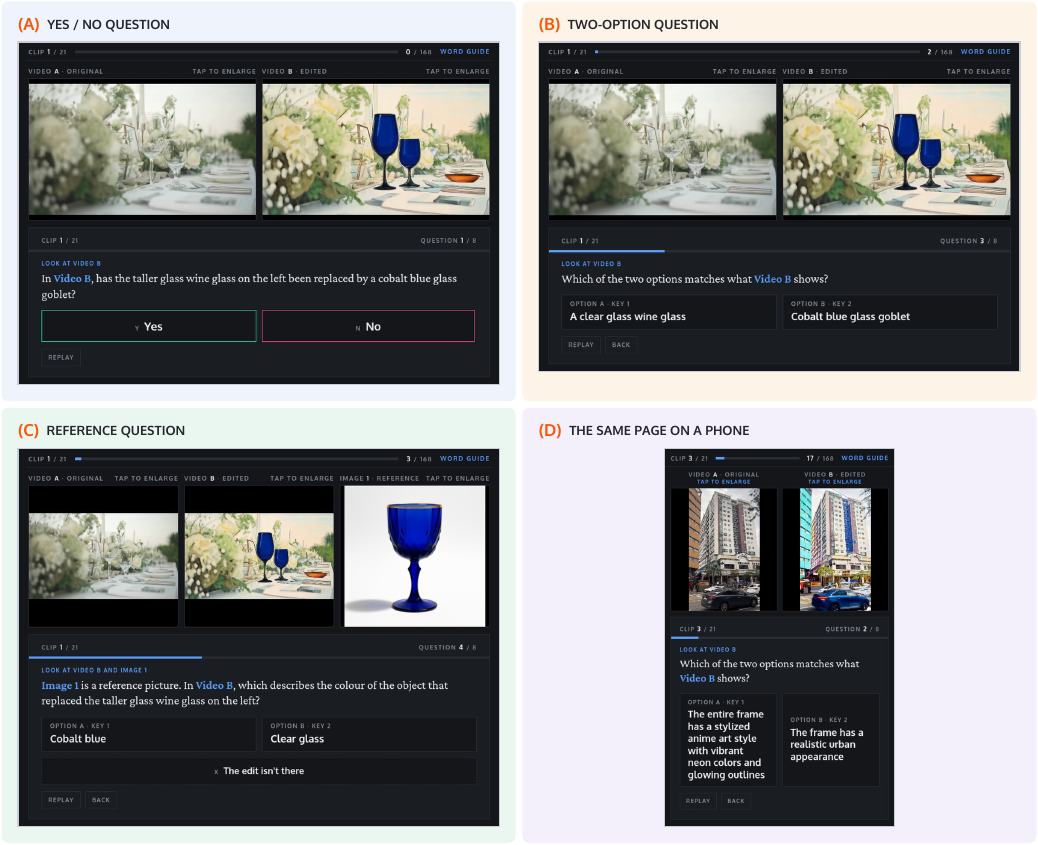}
  \vspace{-12pt}%
  \caption{\textbf{The user study page.} (A)~A yes/no question, (B)~a two-option question, (C)~a reference question with
  the reference image, and (D)~the page on a phone. Video~A is the source and Video~B the edit.}
  \label{fig:app:survey}
\end{suppfloat}

\begin{table}[!ht]
  \centering
  \caption{\textbf{Human scores on the user study.} Percentage of expected answers from people and from the judge, on the
  same 240 questions per method. \emph{Execution} covers the execution and absence questions (153), \emph{Realism} the realism, persistence and
  scope questions (53), and \emph{All} adds the binding and reference questions (17 each). \textbf{Bold} is best. Last row: Spearman's $\rho$
  between the two rankings of the six methods.}
  \label{tab:app:human}
  \supptabfont\setlength{\tabcolsep}{6pt}\renewcommand{\arraystretch}{0.9}
  \begin{tabular}{@{}l*{6}{c}@{}}
    \toprule
    \multirow{2}{*}{Method} & \multicolumn{2}{c}{Execution} & \multicolumn{2}{c}{Realism} & \multicolumn{2}{c}{All} \\
    \cmidrule(lr){2-3}\cmidrule(lr){4-5}\cmidrule(l){6-7}
    & People & Judge & People & Judge & People & Judge \\
    \midrule
    Seedance 2.5~\cite{seedance25} & 90.0 & 93.5 & \textbf{69.5} & \textbf{81.1} & \textbf{82.2} & \textbf{90.0} \\
    CoinVE-Edit~\cite{coinve} & 89.1 & \textbf{94.8} & 63.5 & \textbf{81.1} & 77.8 & 89.6 \\
    LTX-Ripple~\cite{ltxripple} & \textbf{90.5} & 94.1 & 67.7 & 77.4 & 80.9 & 89.6 \\
    \midrule
    \rowcolor{baserow}
    Kiwi-Edit (KE)~\cite{kiwiedit}, joint prompt & 63.0 & 64.7 & 36.9 & 37.7 & 52.5 & 54.2 \\
    \rowcolor{oursrow}
    KE + \textbf{\method} (Ours) & 72.1 & 77.8 & 33.8 & 60.4 & 59.3 & 70.0 \\
    \midrule
    \rowcolor{oursrow}
    JAI + \textbf{\method} (Ours) & 88.4 & 92.8 & 47.0 & 66.0 & 75.0 & 85.8 \\
    \midrule
    Spearman's $\rho$, people and judge & \multicolumn{2}{c}{0.83} & \multicolumn{2}{c}{0.84} & \multicolumn{2}{c}{0.99} \\
    \bottomrule
  \end{tabular}
\end{table}

\suppsec{Additional results}%

\label{sec:app:results}

\subsection{\mt{LND} for every pair of $k$ and $n$}
\label{sec:app:cells}
\mbox{\Cref{fig:teaser}-(D)} pools \mt{LND} over $n$ for each $k$, and over $k$ for each $n$. \Cref{tab:app:cells} gives
every pair of $k$ and $n$. On Kiwi-Edit~\cite{kiwiedit}, \method lands the most edits in all 14 pairs, between 24.2\% and
49.2\%, while the joint prompt falls from 44.4\% at $k = 2$, $n = 1$ to 7.2\% at $k = n = 5$, and sequential
editing from 19.0\% to 0\%. On JoyAI-Video-Edit~\cite{joyai}, \method lands the most edits in 12 of the 14 pairs, and the
joint prompt in the other two, $(k, n) = (4, 3)$ and $(5, 4)$.

\begin{table}[!ht]
  \centering
  \caption{\textbf{\mt{LND} for every pair of $k$ and $n$} on \bench. Each cell pools the instructions of its records, 123 to 128
  instructions over 25 to 63 records. Per backbone, \textbf{bold} is best, \underline{underlined} second.}
  \label{tab:app:cells}
  \supptabfont\setlength{\tabcolsep}{3.7pt}
  \begin{tabular}{@{}l*{14}{c}@{}}
    \toprule
    \multirow{2}{*}{Method} & \multicolumn{2}{c}{$k = 2$} & \multicolumn{3}{c}{$k = 3$} & \multicolumn{4}{c}{$k = 4$} & \multicolumn{5}{c}{$k = 5$} \\
    \cmidrule(lr){2-3}\cmidrule(lr){4-6}\cmidrule(lr){7-10}\cmidrule(l){11-15}
    & $n = 1$ & $n = 2$ & $n = 1$ & $n = 2$ & $n = 3$ & $n = 1$ & $n = 2$ & $n = 3$ & $n = 4$ & $n = 1$ & $n = 2$ & $n = 3$ & $n = 4$ & $n = 5$ \\
    \midrule
    Kiwi-Edit (KE)~\cite{kiwiedit}, sequential & 19.0 & 12.1 & 5.6 & 3.2 & 5.7 & 4.7 & 4.8 & 1.6 & 0.8 & 4.8 & 4.0 & 1.6 & 1.6 & 0.0 \\
    \rowcolor{baserow}
    KE, joint prompt & \underline{44.4} & \underline{36.3} & \underline{31.0} & \underline{27.8} & \underline{29.3} & \underline{25.0} & \underline{27.4} & \underline{12.9} & \underline{14.5} & \underline{38.4} & \underline{24.8} & \underline{12.0} & \underline{12.0} & \underline{7.2} \\
    \rowcolor{oursrow}
    KE + \textbf{\method} (Ours) & \textbf{46.0} & \textbf{49.2} & \textbf{36.5} & \textbf{35.7} & \textbf{48.8} & \textbf{34.4} & \textbf{37.9} & \textbf{24.2} & \textbf{33.9} & \textbf{40.0} & \textbf{40.0} & \textbf{27.2} & \textbf{31.2} & \textbf{36.0} \\
    \midrule
    JoyAI (JAI)~\cite{joyai}, sequential & 47.6 & 40.3 & 21.4 & 31.7 & 37.4 & 26.6 & 27.4 & 26.6 & 26.6 & 29.6 & 26.4 & 19.2 & 20.0 & 18.4 \\
    \rowcolor{baserow}
    JAI, joint prompt & \underline{58.7} & \underline{59.7} & \underline{49.2} & \underline{54.0} & \underline{64.2} & \underline{56.2} & \underline{44.4} & \textbf{51.6} & \underline{59.7} & \underline{48.0} & \underline{48.8} & \underline{50.4} & \textbf{56.0} & \underline{56.8} \\
    \rowcolor{oursrow}
    JAI + \textbf{\method} (Ours) & \textbf{65.1} & \textbf{65.3} & \textbf{55.6} & \textbf{57.1} & \textbf{66.7} & \textbf{58.6} & \textbf{46.0} & \underline{46.0} & \textbf{65.3} & \textbf{53.6} & \textbf{52.0} & \textbf{58.4} & \underline{50.4} & \textbf{63.2} \\
    \bottomrule
  \end{tabular}
\end{table}

\suppnobreak
\subsection{Robustness to instruction order}
\label{sec:app:order}
A joint prompt lists the $k$ instructions in some order, and \method concatenates their contexts in the same order. This
section tests whether that order matters. We ask whether \method gives the same result when the order changes, and whether
it still improves its backbone.

\paragraph{Setup.} We use all 500 records of \bench. In each record we reverse the order of the instructions, keep each
reference with its instruction, and change nothing else. We run the joint prompt and \method on Kiwi-Edit and on
JoyAI-Video-Edit in both orders, with the settings of \cref{tab:app:hparams} and the same checklist questions.
\Cref{tab:app:order} reports the four rows of each backbone with the metrics of \cref{tab:main}.

\paragraph{Effect of the order.} \method is robust to the order. With the reversed order, \mt{LND} moves from 37.20 to 38.23
for KE + \method and from 57.37 to 57.71 for JAI + \method, and \mt{ALL} from 10.00 to 10.00 and from 23.00 to 22.20,
respectively. \method encodes each instruction alone and assigns it to its own region, so no instruction depends on its
position in a prompt. The joint prompt is also stable: \mt{LND} moves from 24.51 to 24.69 on Kiwi-Edit and from 54.11 to
54.80 on JoyAI-Video-Edit, respectively. Its failed edits therefore do not come from the order of the instructions.

\paragraph{Gain over the joint prompt.} \method lands more edits than the joint prompt in both orders. On Kiwi-Edit,
\mt{LND} rises by 12.69 points in the original order and 13.54 in the reversed order, and on JoyAI-Video-Edit by 3.26 and
2.91. \mt{ALL} rises in both orders: from 5.40 and 5.80 to 10.00 on Kiwi-Edit, and from 20.80 and
20.20 to 23.00 and 22.20 on JoyAI-Video-Edit.

\begin{table}[!ht]
  \centering
  \caption{\textbf{Robustness to instruction order} on \bench, with the metrics of \cref{tab:main}. Each method runs on all 500
  records with the instructions in their original and in reversed order (each reference stays with its instruction). Per backbone, \textbf{bold} is best, \underline{underlined} second.}
  \label{tab:app:order}
  \supptabfont\setlength{\tabcolsep}{2.9pt}
  \begin{tabular}{@{}ll*{15}{c}@{}}
    \toprule
    \multirow{2}{*}{Method} & \multirow{2}{*}{Order} & \multicolumn{3}{c}{Instr.\ compliance} & \multicolumn{2}{c}{Ref.\ fidelity} & \multicolumn{2}{>{\columncolor{succN}}c}{\textbf{Success}} & \multicolumn{4}{c}{Video fidelity} & \multicolumn{4}{c}{Video quality} \\
    \cmidrule(lr){3-5}\cmidrule(lr){6-7}\cmidrule(lr){8-9}\cmidrule(lr){10-13}\cmidrule(l){14-17}
    & & \mt{EXE} & \mt{EQ} & \mt{BND} & \mt{RA} & \mt{RS} & \cellcolor{succN}\textbf{\mt{LND}} & \cellcolor{succN}\textbf{\mt{ALL}} & \mt{SEM} & \mt{SRC} & \mt{SSO} & \mt{MF} & \mt{VQR} & \mt{AES} & \mt{TQ} & \mt{MSM} \\
    \midrule
    \multicolumn{17}{@{}l}{\textbf{\textit{Kiwi-Edit (KE)}}~\cite{kiwiedit}} \\
    \rowcolor{baserow}
    KE, joint prompt & original  & 61.27 & 42.68 & 34.46 & 32.27 & 0.617 & \cellcolor{succB}24.51 & \cellcolor{succB}5.40 & 7.11 & 0.794 & \underline{0.531} & 0.910 & 3.86 & 4.80 & 15.12 & \underline{0.756} \\
    \rowcolor{baserow}
    KE, joint prompt & reversed & 60.35 & 42.98 & 34.07 & 32.74 & 0.613 & \cellcolor{succB}24.69 & \cellcolor{succB}\underline{5.80} & 7.27 & \underline{0.799} & \textbf{0.537} & \underline{0.916} & 3.87 & 4.82 & 15.03 & \textbf{0.757} \\
    \rowcolor{oursrow}
    KE + \textbf{\method} (Ours) & original  & \textbf{74.60} & \underline{56.27} & \underline{59.77} & \textbf{47.26} & \textbf{0.619} & \cellcolor{succO}\underline{37.20} & \cellcolor{succO}\textbf{10.00} & \underline{7.86} & \textbf{0.805} & 0.505 & \textbf{0.937} & \underline{4.11} & \textbf{4.96} & \underline{15.98} & 0.753 \\
    \rowcolor{oursrow}
    KE + \textbf{\method} (Ours) & reversed & \underline{74.41} & \textbf{57.02} & \textbf{60.44} & \underline{47.20} & \underline{0.618} & \cellcolor{succO}\textbf{38.23} & \cellcolor{succO}\textbf{10.00} & \textbf{7.89} & \textbf{0.805} & 0.506 & \textbf{0.937} & \textbf{4.12} & \underline{4.95} & \textbf{15.99} & 0.753 \\
    \midrule
    \multicolumn{17}{@{}l}{\textbf{\textit{JoyAI-Video-Edit (JAI)}}~\cite{joyai}} \\
    \rowcolor{baserow}
    JAI, joint prompt & original  & 87.26 & 72.96 & 77.38 & 64.99 & 0.635 & \cellcolor{succB}54.11 & \cellcolor{succB}20.80 & 9.26 & \underline{0.838} & 0.740 & 0.946 & \underline{3.65} & \textbf{4.44} & \textbf{13.81} & 0.742 \\
    \rowcolor{baserow}
    JAI, joint prompt & reversed & 87.74 & 73.49 & 77.00 & 64.19 & 0.633 & \cellcolor{succB}54.80 & \cellcolor{succB}20.20 & 9.30 & 0.837 & \underline{0.746} & 0.948 & 3.62 & \textbf{4.44} & \underline{13.76} & \underline{0.743} \\
    \rowcolor{oursrow}
    JAI + \textbf{\method} (Ours) & original  & \underline{88.15} & \underline{75.18} & \underline{81.23} & \textbf{66.69} & \textbf{0.638} & \cellcolor{succO}\underline{57.37} & \cellcolor{succO}\textbf{23.00} & \underline{9.37} & \textbf{0.841} & \textbf{0.783} & \underline{0.949} & \textbf{3.66} & \underline{4.39} & 13.71 & \textbf{0.747} \\
    \rowcolor{oursrow}
    JAI + \textbf{\method} (Ours) & reversed & \textbf{88.41} & \textbf{75.84} & \textbf{81.42} & \underline{66.32} & \underline{0.636} & \cellcolor{succO}\textbf{57.71} & \cellcolor{succO}\underline{22.20} & \textbf{9.39} & \textbf{0.841} & \textbf{0.783} & \textbf{0.950} & \underline{3.65} & \underline{4.39} & 13.74 & \textbf{0.747} \\
    \bottomrule
  \end{tabular}
\end{table}

\suppnobreak
\subsection{Attention masks against SAM~3.1 masks}
\label{sec:app:sam}
\textit{Locate} takes its masks from the cross-attention of the editor. A segmenter is the usual alternative, since it gives
sharp masks of the objects in the source video. This section tests whether \method lands more edits with the masks of a
segmenter, SAM~3.1~\cite{sam3}, than with our attention masks.

\paragraph{Setup.} We replace only the masks of \method and keep every other setting. SAM~3.1 takes one noun per instruction
and tracks the edited object on the source video. Where SAM~3.1 finds nothing, as for every added object, which is not in the
source video, each backbone keeps its own attention mask (707 of the 1{,}750 instructions, 40\%). \Cref{tab:app:sam} therefore
reports all 500 records and also the 108 records where SAM~3.1 gives every mask.

\paragraph{Kiwi-Edit.}
SAM~3.1 masks land about as many edits as our masks. \mt{LND} is 37.20 with our masks and 38.46 with SAM~3.1 masks on all
records, and 39.94 and 40.58 on the 108 records, respectively. \mt{ALL} is 10.00 and 9.20 on all records, and 13.89 and 12.96
on the 108 records, respectively. The difference in \mt{LND} is within the variation between records: 155 instructions land
only with SAM~3.1 masks and 133 only with ours (sign test, $p = 0.22$). SAM~3.1 masks leak more onto the other targets:
\mt{BND} falls from 59.77 to 57.56 on all records and from 67.01 to 59.79 on the 108 records. They also keep slightly less of
the untouched video (\mt{SSO} 0.505 against 0.496).

\paragraph{JoyAI-Video-Edit.}
JAI + \method lands slightly fewer edits with SAM~3.1 masks. \mt{LND} is 57.37 with our masks and 55.66 with SAM~3.1 masks
on all records, and 55.84 and 53.57 on the 108 records, respectively. \mt{ALL} is 23.00 and 21.20 on all records, and 25.00
and 20.37 on the 108 records, respectively. 186 instructions land only with our masks and 156 only with SAM~3.1 masks
($p = 0.12$). The drop is largest on background edits. SAM~3.1 has no object to track there, so their mask covers
everything outside the other regions, and \mt{LND} falls from 55.7 to 49.6 on these 115 instructions. Where SAM~3.1 tracks
the edited object by its noun, \mt{LND} changes little (56.0 to 54.6). As on Kiwi-Edit, SAM~3.1 masks leak more
onto the other targets (\mt{BND} 81.23 to 76.52).

\paragraph{Summary.}
On neither backbone do SAM~3.1 masks land clearly more edits than our attention masks, and on both they leak more onto the
other targets. Our masks come from the attention of the editor itself, so they need neither a segmenter nor a noun for each
instruction.

\begin{table}[!ht]
  \centering
  \caption{\textbf{Attention masks against SAM~3.1 masks} on \bench, with the metrics of \cref{tab:main}. Each pair differs
  only in the masks: our attention masks (green) or SAM~3.1~\cite{sam3} masks (\cref{sec:app:sam}). Top, all 500 records; bottom,
  the 108 records where SAM~3.1 finds every object. In each pair, \textbf{bold} is better.}
  \label{tab:app:sam}
  \supptabfont\setlength{\tabcolsep}{3.3pt}
  \begin{tabular}{@{}l*{15}{c}@{}}
    \toprule
    \multirow{2}{*}{Variant} & \multicolumn{3}{c}{Instr.\ compliance} & \multicolumn{2}{c}{Ref.\ fidelity} & \multicolumn{2}{>{\columncolor{succN}}c}{\textbf{Success}} & \multicolumn{4}{c}{Video fidelity} & \multicolumn{4}{c}{Video quality} \\
    \cmidrule(lr){2-4}\cmidrule(lr){5-6}\cmidrule(lr){7-8}\cmidrule(lr){9-12}\cmidrule(l){13-16}
    & \mt{EXE} & \mt{EQ} & \mt{BND} & \mt{RA} & \mt{RS} & \cellcolor{succN}\textbf{\mt{LND}} & \cellcolor{succN}\textbf{\mt{ALL}} & \mt{SEM} & \mt{SRC} & \mt{SSO} & \mt{MF} & \mt{VQR} & \mt{AES} & \mt{TQ} & \mt{MSM} \\
    \midrule
    \multicolumn{16}{@{}l}{\textbf{\textit{All 500 records}}} \\
    \rowcolor{oursrow}
    KE + \textbf{\method}, our masks & 74.60 & 56.27 & \textbf{59.77} & 47.26 & 0.619 & \cellcolor{succO}37.20 & \cellcolor{succO}\textbf{10.00} & \textbf{7.86} & 0.805 & \textbf{0.505} & \textbf{0.937} & \textbf{4.11} & \textbf{4.96} & \textbf{15.98} & \textbf{0.753} \\
    KE + \textbf{\method}, SAM~3.1 masks & \textbf{74.73} & \textbf{57.19} & 57.56 & \textbf{47.85} & \textbf{0.633} & \cellcolor{succN}\textbf{38.46} & \cellcolor{succN}9.20 & 7.84 & \textbf{0.806} & 0.496 & 0.934 & \textbf{4.11} & 4.95 & 15.93 & 0.751 \\
    \cmidrule(lr){1-16}
    \rowcolor{oursrow}
    JAI + \textbf{\method}, our masks & \textbf{88.15} & \textbf{75.18} & \textbf{81.23} & \textbf{66.69} & 0.638 & \cellcolor{succO}\textbf{57.37} & \cellcolor{succO}\textbf{23.00} & \textbf{9.37} & 0.841 & 0.783 & 0.949 & \textbf{3.66} & \textbf{4.39} & 13.71 & \textbf{0.747} \\
    JAI + \textbf{\method}, SAM~3.1 masks & 85.01 & 73.67 & 76.52 & 64.21 & \textbf{0.640} & \cellcolor{succN}55.66 & \cellcolor{succN}21.20 & 9.36 & \textbf{0.842} & \textbf{0.785} & \textbf{0.951} & 3.65 & \textbf{4.39} & \textbf{13.72} & \textbf{0.747} \\
    \midrule
    \multicolumn{16}{@{}l}{\textbf{\textit{108 records, every mask from SAM~3.1}}} \\
    \rowcolor{oursrow}
    KE + \textbf{\method}, our masks & \textbf{78.61} & 61.09 & \textbf{67.01} & 55.24 & 0.604 & \cellcolor{succO}39.94 & \cellcolor{succO}\textbf{13.89} & \textbf{8.00} & \textbf{0.811} & \textbf{0.456} & \textbf{0.934} & \textbf{4.11} & 4.91 & \textbf{16.13} & 0.751 \\
    KE + \textbf{\method}, SAM~3.1 masks & 78.38 & \textbf{63.15} & 59.79 & \textbf{58.24} & \textbf{0.623} & \cellcolor{succN}\textbf{40.58} & \cellcolor{succN}12.96 & 7.86 & \textbf{0.811} & 0.452 & 0.927 & 4.07 & \textbf{4.92} & 15.86 & \textbf{0.752} \\
    \cmidrule(lr){1-16}
    \rowcolor{oursrow}
    JAI + \textbf{\method}, our masks & \textbf{87.60} & \textbf{75.27} & \textbf{82.47} & 65.36 & \textbf{0.625} & \cellcolor{succO}\textbf{55.84} & \cellcolor{succO}\textbf{25.00} & \textbf{9.44} & 0.854 & 0.755 & 0.944 & \textbf{3.67} & 4.37 & 14.01 & \textbf{0.750} \\
    JAI + \textbf{\method}, SAM~3.1 masks & 85.55 & 73.58 & 76.29 & \textbf{66.29} & 0.623 & \cellcolor{succN}53.57 & \cellcolor{succN}20.37 & 9.37 & \textbf{0.856} & \textbf{0.759} & \textbf{0.947} & 3.64 & \textbf{4.39} & \textbf{14.13} & 0.748 \\
    \bottomrule
  \end{tabular}
\end{table}

\subsection{Upper bound for referenced instructions}
\label{sec:app:rsbound}
\method keeps each backbone frozen, so the backbone limits how well a referenced edit can turn out. This section measures
that limit: how well each backbone performs a referenced instruction when the instruction is alone. We then ask how much of
the limit the joint prompt loses, and how much \method recovers, when all $k$ instructions are edited at once.

\paragraph{Setup.} We take the 1{,}122 instructions of \bench that have a reference. For the upper bound, we render each of
them alone from the source video with its reference, on Kiwi-Edit and on JoyAI-Video-Edit, with the backbone settings of
\cref{tab:app:hparams}. No other instruction or reference competes with it. The upper bound is not a method: it gives one
video per instruction, each with a single edit. The judge answers only the checklist questions of that instruction. The
joint prompt and \method render each record once with all $k$ instructions. \Cref{tab:app:rsbound} reports the three over
the same 1{,}122 instructions.

\paragraph{Landed edits.} Alone, Kiwi-Edit and JoyAI-Video-Edit land 56.95\% and 74.51\% of the instructions, respectively.
The joint prompt lands 21.30\% and 54.90\%, so it loses 35.65 and 19.61 points of \mt{LND}, respectively, to the competition
between instructions. \method recovers 13.01 and 3.92 of these points, respectively, about a third and a fifth of the loss,
without training. \mt{RA} shows the same order: 70.19 alone, 32.27 with the joint prompt and 47.26 with \method on
Kiwi-Edit, and 77.07, 64.99 and 66.69, respectively, on JoyAI-Video-Edit.

\paragraph{Reference similarity.} \mt{RS} has less room than \mt{LND}. Alone, Kiwi-Edit and JoyAI-Video-Edit reach 0.663 and
0.664, respectively, against 0.617 and 0.635 with the joint prompt and 0.619 and 0.638 with \method. \mt{RS} is averaged only
over the instructions whose edited object is found in the result, and this set grows with \method: from 439 to 617
instructions on Kiwi-Edit and from 766 to 781 on JoyAI-Video-Edit, against 774 and 867 alone, respectively. \method therefore
keeps \mt{RS} at the backbone's level while more referenced edits appear.

\begin{table}[!ht]
  \centering
  \caption{\textbf{Upper bound for referenced instructions on \bench.} ``One instruction alone'' is the upper bound of each backbone: every
  referenced instruction is rendered alone with its reference. The joint prompt and \method edit all $k$ instructions of a record at once. All columns
  are over the 1{,}122 instructions with a reference. Found: the instructions whose edited object is detected in the result, as a count and as a share of the 1{,}122; \mt{RS} is averaged over them. Per backbone, \textbf{bold} is best, \underline{underlined} second.}
  \label{tab:app:rsbound}
  \supptabfont\setlength{\tabcolsep}{5pt}
  \begin{tabular}{@{}l*{7}{c}@{}}
    \toprule
    \multirow{2}{*}{Method} & \multicolumn{3}{c}{Instr.\ compliance} & \multicolumn{3}{c}{Ref.\ fidelity} & \multicolumn{1}{>{\columncolor{succN}}c}{\textbf{Success}} \\
    \cmidrule(lr){2-4}\cmidrule(lr){5-7}\cmidrule(l){8-8}
    & \mt{EXE} & \mt{EQ} & \mt{BND} & \mt{RA} & Found & \mt{RS} & \cellcolor{succN}\textbf{\mt{LND}} \\
    \midrule
    Kiwi-Edit (KE)~\cite{kiwiedit}, one instruction alone & \textbf{91.31} & \textbf{75.00} & \textbf{80.42} & \textbf{70.19} & \textbf{774 (69.0\%)} & \textbf{0.663} & \cellcolor{succN}\textbf{56.95} \\
    \rowcolor{baserow}
    KE, joint prompt & 61.31 & 41.36 & 34.86 & 32.27 & 439 (39.1\%) & 0.617 & \cellcolor{succB}21.30 \\
    \rowcolor{oursrow}
    KE + \textbf{\method} (Ours) & \underline{75.30} & \underline{56.09} & \underline{59.17} & \underline{47.26} & \underline{617 (55.0\%)} & \underline{0.619} & \cellcolor{succO}\underline{34.31} \\
    \midrule
    JoyAI (JAI)~\cite{joyai}, one instruction alone & \textbf{96.96} & \textbf{87.22} & \textbf{92.64} & \textbf{77.07} & \textbf{867 (77.3\%)} & \textbf{0.664} & \cellcolor{succN}\textbf{74.51} \\
    \rowcolor{baserow}
    JAI, joint prompt & 89.67 & 75.03 & 76.53 & 64.99 & 766 (68.3\%) & 0.635 & \cellcolor{succB}54.90 \\
    \rowcolor{oursrow}
    JAI + \textbf{\method} (Ours) & \underline{91.52} & \underline{77.68} & \underline{80.97} & \underline{66.69} & \underline{781 (69.6\%)} & \underline{0.638} & \cellcolor{succO}\underline{58.82} \\
    \bottomrule
  \end{tabular}
\end{table}

\subsection{More ablations}
\label{sec:app:xroute}
\label{sec:app:ladders}
We ablate the design choices of \method on Kiwi-Edit over all 500 records. Each of \cref{tab:abl:step,tab:abl:blocks,tab:abl:overlap,tab:abl:rule,tab:abl:etar,tab:abl:iblocks,tab:abl:etai,tab:abl:boost}
changes one setting and keeps every other setting of the full method, whose row is repeated in green. The first three vary how
the masks are extracted (\cref{sec:ours:locate}), the next two the reference assignment in self-attention, and the last three
the instruction assignment in cross-attention (\cref{sec:ours:render}).

\paragraph{Mask step $s$.} \textit{Locate} reads the masks after $s$ steps (\cref{sec:ours:locate}). \Cref{tab:abl:step}
compares $s = 3$, 5, 8 and 10. \method is not sensitive to $s$: across these steps, \mt{LND} stays between 37.09 and 38.34 and \mt{ALL} between 9.00 and 10.00. An early step lands
slightly more instructions (\mt{LND} 38.34 at $s = 3$), and a late step keeps slightly more of the untouched video (\mt{SSO}
0.500 at $s = 3$ against 0.510 at $s = 8$ and 10). A larger $s$ also costs more, since \textit{Locate} runs $s$ steps for
every instruction. We use $s = 5$, which has the highest \mt{ALL} (10.00).

\begin{table}[!ht]
  \centering
  \caption{\textbf{Mask step $s$.} KE + \method on all 500 records of \bench, with the metrics of \cref{tab:main}; the row of \method is green. \textbf{Bold} is best, \underline{underlined} second.}
  \label{tab:abl:step}
  \supptabfont\setlength{\tabcolsep}{3.3pt}\renewcommand{\arraystretch}{0.9}
  \begin{tabular}{@{}l*{15}{c}@{}}
    \toprule
    \multirow{2}{*}{Variant} & \multicolumn{3}{c}{Instr.\ compliance} & \multicolumn{2}{c}{Ref.\ fidelity} & \multicolumn{2}{>{\columncolor{succN}}c}{\textbf{Success}} & \multicolumn{4}{c}{Video fidelity} & \multicolumn{4}{c}{Video quality} \\
    \cmidrule(lr){2-4}\cmidrule(lr){5-6}\cmidrule(lr){7-8}\cmidrule(lr){9-12}\cmidrule(l){13-16}
    & \mt{EXE} & \mt{EQ} & \mt{BND} & \mt{RA} & \mt{RS} & \cellcolor{succN}\textbf{\mt{LND}} & \cellcolor{succN}\textbf{\mt{ALL}} & \mt{SEM} & \mt{SRC} & \mt{SSO} & \mt{MF} & \mt{VQR} & \mt{AES} & \mt{TQ} & \mt{MSM} \\
    \midrule
    $s = 3$ & \underline{74.38} & \textbf{57.00} & 59.67 & \underline{47.58} & 0.619 & \cellcolor{succN}\textbf{38.34} & \cellcolor{succN}9.60 & \underline{7.86} & 0.804 & 0.500 & \underline{0.935} & \textbf{4.12} & \textbf{4.97} & \textbf{15.99} & \textbf{0.754} \\
    \rowcolor{oursrow}
    $s = 5$ (\textbf{\method}) & \textbf{74.60} & \underline{56.27} & 59.77 & 47.26 & 0.619 & \cellcolor{succO}37.20 & \cellcolor{succO}\textbf{10.00} & \underline{7.86} & \underline{0.805} & \underline{0.505} & \textbf{0.937} & \underline{4.11} & \underline{4.96} & \underline{15.98} & \underline{0.753} \\
    $s = 8$ & 74.32 & \textbf{57.00} & \textbf{60.64} & \textbf{48.69} & 0.619 & \cellcolor{succN}\underline{37.31} & \cellcolor{succN}\underline{9.80} & \textbf{7.94} & \underline{0.805} & \textbf{0.510} & \textbf{0.937} & \textbf{4.12} & \underline{4.96} & 15.97 & \underline{0.753} \\
    $s = 10$ & 73.79 & 55.94 & \underline{60.15} & 47.35 & 0.619 & \cellcolor{succN}37.09 & \cellcolor{succN}9.00 & \underline{7.86} & \textbf{0.806} & \textbf{0.510} & \textbf{0.937} & \underline{4.11} & \underline{4.96} & 15.94 & \underline{0.753} \\
    \bottomrule
  \end{tabular}
\end{table}

\paragraph{Mask blocks $L$.} \textit{Locate} averages the cross-attention maps of the $L$ blocks with the highest $\delta_l$
(\cref{sec:ours:locate}). \Cref{tab:abl:blocks} compares $L = 10$ with $L = 20$ and with all 30 blocks. \method is not
sensitive to $L$. \mt{LND} is 37.20, 38.29 and 36.91 with $L = 10$, 20 and 30, and \mt{ALL} is 10.00, 12.00 and 9.80, so
neither score rises or falls steadily with $L$. $L = 20$ scores highest, but its gain over $L = 10$ is not significant in a
paired sign test: 108 instructions land only with $L = 20$ and 89 only with $L = 10$ ($p = 0.20$), and 23 records land in
full only with $L = 20$ and 13 only with $L = 10$ ($p = 0.13$). $L = 30$ does not differ significantly from $L = 10$
either: 107 instructions land only with $L = 30$ and 112 only with $L = 10$ ($p = 0.79$), and 15 records land in full only
with $L = 30$ and 16 only with $L = 10$ ($p = 1.00$). We fixed $L = 10$ before this ablation and use $L = 10$ in all
experiments, so $L$ is not tuned on \bench.

\begin{table}[!ht]
  \centering
  \caption{\textbf{Mask blocks $L$,} as in \cref{tab:abl:step}. \textbf{Bold} is best, \underline{underlined} second.}
  \label{tab:abl:blocks}
  \supptabfont\setlength{\tabcolsep}{3.3pt}\renewcommand{\arraystretch}{0.9}
  \begin{tabular}{@{}l*{15}{c}@{}}
    \toprule
    \multirow{2}{*}{Variant} & \multicolumn{3}{c}{Instr.\ compliance} & \multicolumn{2}{c}{Ref.\ fidelity} & \multicolumn{2}{>{\columncolor{succN}}c}{\textbf{Success}} & \multicolumn{4}{c}{Video fidelity} & \multicolumn{4}{c}{Video quality} \\
    \cmidrule(lr){2-4}\cmidrule(lr){5-6}\cmidrule(lr){7-8}\cmidrule(lr){9-12}\cmidrule(l){13-16}
    & \mt{EXE} & \mt{EQ} & \mt{BND} & \mt{RA} & \mt{RS} & \cellcolor{succN}\textbf{\mt{LND}} & \cellcolor{succN}\textbf{\mt{ALL}} & \mt{SEM} & \mt{SRC} & \mt{SSO} & \mt{MF} & \mt{VQR} & \mt{AES} & \mt{TQ} & \mt{MSM} \\
    \midrule
    \rowcolor{oursrow}
    $L = 10$ (\textbf{\method}) & \underline{74.60} & 56.27 & 59.77 & \underline{47.26} & \textbf{0.619} & \cellcolor{succO}\underline{37.20} & \cellcolor{succO}\underline{10.00} & \underline{7.86} & \underline{0.805} & \underline{0.505} & \textbf{0.937} & \underline{4.11} & 4.96 & \underline{15.98} & \underline{0.753} \\
    $L = 20$ & 74.34 & \textbf{57.59} & \textbf{61.12} & \textbf{48.42} & \underline{0.617} & \cellcolor{succN}\textbf{38.29} & \cellcolor{succN}\textbf{12.00} & 7.85 & \underline{0.805} & \textbf{0.508} & \textbf{0.937} & \textbf{4.12} & 4.96 & \textbf{16.02} & \textbf{0.754} \\
    $L = 30$ (all blocks) & \textbf{74.71} & \underline{57.04} & \underline{60.64} & 47.22 & \underline{0.617} & \cellcolor{succN}36.91 & \cellcolor{succN}9.80 & \textbf{7.88} & \textbf{0.806} & \textbf{0.508} & \underline{0.936} & \textbf{4.12} & 4.96 & \textbf{16.02} & \underline{0.753} \\
    \bottomrule
  \end{tabular}
\end{table}

\paragraph{Mask overlap.} \textit{Locate} keeps each video token in at most one mask: a token in several candidate sets
$\tilde{M}_i$ goes to the instruction with the largest contrast (\cref{eq:mask}). We drop this step and keep the candidate
sets as they are, so a token in several sets attends to each of their instructions and references. \Cref{tab:abl:overlap}
compares these overlapping masks with the disjoint masks of \method. 338 of the 500 records
have such tokens, 6.3\% of the masked tokens.
\method is not sensitive to this choice. With overlapping masks, \mt{LND} is 36.86 against 37.20 and \mt{ALL} is 11.00
against 10.00. 99 instructions land only with overlapping masks and 105 only with disjoint masks (sign test, $p = 0.73$), and
17 records land in full only with overlapping masks and 12 only with disjoint masks ($p = 0.46$). \mt{BND} falls from 59.77 to
58.42, which is also within the variation between records ($p = 0.20$). \mt{SEM} is 7.79 against 7.86, and the other video
metrics change by at most 0.02. We keep the masks disjoint, so that each video token follows exactly
one instruction and one reference.

\begin{table}[!ht]
  \centering
  \caption{\textbf{Mask overlap,} as in \cref{tab:abl:step}. \textbf{Bold} is best, \underline{underlined} second.}
  \label{tab:abl:overlap}
  \supptabfont\setlength{\tabcolsep}{3.3pt}\renewcommand{\arraystretch}{0.9}
  \begin{tabular}{@{}l*{15}{c}@{}}
    \toprule
    \multirow{2}{*}{Variant} & \multicolumn{3}{c}{Instr.\ compliance} & \multicolumn{2}{c}{Ref.\ fidelity} & \multicolumn{2}{>{\columncolor{succN}}c}{\textbf{Success}} & \multicolumn{4}{c}{Video fidelity} & \multicolumn{4}{c}{Video quality} \\
    \cmidrule(lr){2-4}\cmidrule(lr){5-6}\cmidrule(lr){7-8}\cmidrule(lr){9-12}\cmidrule(l){13-16}
    & \mt{EXE} & \mt{EQ} & \mt{BND} & \mt{RA} & \mt{RS} & \cellcolor{succN}\textbf{\mt{LND}} & \cellcolor{succN}\textbf{\mt{ALL}} & \mt{SEM} & \mt{SRC} & \mt{SSO} & \mt{MF} & \mt{VQR} & \mt{AES} & \mt{TQ} & \mt{MSM} \\
    \midrule
    Masks may overlap & \underline{74.45} & \textbf{56.51} & \underline{58.42} & \underline{46.80} & \underline{0.618} & \cellcolor{succN}\underline{36.86} & \cellcolor{succN}\textbf{11.00} & \underline{7.79} & 0.805 & \textbf{0.506} & \underline{0.936} & \textbf{4.12} & 4.96 & \textbf{16.00} & 0.753 \\
    \rowcolor{oursrow}
    Disjoint masks (\textbf{\method}) & \textbf{74.60} & \underline{56.27} & \textbf{59.77} & \textbf{47.26} & \textbf{0.619} & \cellcolor{succO}\textbf{37.20} & \cellcolor{succO}\underline{10.00} & \textbf{7.86} & 0.805 & \underline{0.505} & \textbf{0.937} & \underline{4.11} & 4.96 & \underline{15.98} & 0.753 \\
    \bottomrule
  \end{tabular}
\end{table}

\paragraph{Reference assignment rule.} \Cref{tab:abl:rules} lists the three restrictions on self-attention, which are the
three cases of \cref{eq:route}, and \cref{tab:abl:rule} applies them in different combinations. \method uses all three.
Every variant keeps instruction assignment in cross-attention. Without reference assignment, every video token sees every
reference: \mt{LND} falls from 37.20 to 33.60 and \mt{ALL} from 10.00 to 7.60, and less of the untouched video is kept
(\mt{SSO} 0.438 against 0.505, \mt{SEM} 6.99 against 7.86). Restriction (a) gives most of the gain: when a masked video
token sees only its own reference, \mt{LND} is 37.26 and \mt{ALL} is 10.20, but less of the untouched video is kept
(\mt{SSO} 0.476, \mt{SEM} 7.58). When the references are also separated from each other (b), \mt{ALL} falls to 6.40: 26
records land in full only with \method and 8 only with this variant (sign test, $p = 0.003$). When the padding is also
hidden from the video tokens (c), the untouched video is restored (\mt{SSO} 0.505, \mt{SEM} 7.89) and \mt{ALL} is 8.20
($p = 0.16$). \method applies (a) to (c) together and keeps both the landing of (a) alone and the untouched video.

\begin{table}[!ht]
  \centering
  \caption{\textbf{Restrictions of the reference assignment rule.} \method uses all three, the three cases of \cref{eq:route}.}
  \label{tab:abl:rules}
  \supptabfont\renewcommand{\arraystretch}{0.9}
  \begin{tabular}{@{}cl@{}}
    \toprule
    & Restriction \\
    \midrule
    (a) & A masked video token sees only its own reference \\
    (b) & References do not see each other \\
    (c) & Video tokens do not see the reference padding \\
    \bottomrule
  \end{tabular}
\end{table}

\begin{table}[!ht]
  \centering
  \caption{\textbf{Reference assignment rule,} as in \cref{tab:abl:step}. \cmark\ marks a restriction of \cref{tab:abl:rules} that is on. \textbf{Bold} is best, \underline{underlined} second.}
  \label{tab:abl:rule}
  \supptabfont\setlength{\tabcolsep}{2.6pt}\renewcommand{\arraystretch}{0.9}
  \begin{tabular}{@{}l*{3}{c}*{15}{c}@{}}
    \toprule
    \multirow{2}{*}{Variant} & \multicolumn{3}{c}{Restriction} & \multicolumn{3}{c}{Instr.\ compliance} & \multicolumn{2}{c}{Ref.\ fidelity} & \multicolumn{2}{>{\columncolor{succN}}c}{\textbf{Success}} & \multicolumn{4}{c}{Video fidelity} & \multicolumn{4}{c}{Video quality} \\
    \cmidrule(lr){2-4}\cmidrule(lr){5-7}\cmidrule(lr){8-9}\cmidrule(lr){10-11}\cmidrule(lr){12-15}\cmidrule(l){16-19}
    & (a) & (b) & (c) & \mt{EXE} & \mt{EQ} & \mt{BND} & \mt{RA} & \mt{RS} & \cellcolor{succN}\textbf{\mt{LND}} & \cellcolor{succN}\textbf{\mt{ALL}} & \mt{SEM} & \mt{SRC} & \mt{SSO} & \mt{MF} & \mt{VQR} & \mt{AES} & \mt{TQ} & \mt{MSM} \\
    \midrule
    No reference assignment & \xmark & \xmark & \xmark & \underline{75.65} & 53.59 & 58.81 & \textbf{47.54} & 0.617 & \cellcolor{succN}33.60 & \cellcolor{succN}7.60 & 6.99 & 0.784 & 0.438 & 0.921 & 4.02 & 4.90 & 15.04 & \textbf{0.753} \\
    Own reference only & \cmark & \xmark & \xmark & \textbf{76.39} & \textbf{57.11} & \textbf{61.69} & \underline{47.28} & \underline{0.618} & \cellcolor{succN}\textbf{37.26} & \cellcolor{succN}\textbf{10.20} & 7.58 & 0.797 & 0.476 & 0.933 & 4.10 & \underline{4.94} & 15.66 & \underline{0.752} \\
    + references separated & \cmark & \cmark & \xmark & 75.21 & 55.59 & \underline{60.25} & 47.03 & \textbf{0.619} & \cellcolor{succN}35.89 & \cellcolor{succN}6.40 & 7.58 & \underline{0.798} & \underline{0.478} & 0.935 & \underline{4.11} & \underline{4.94} & 15.70 & \textbf{0.753} \\
    + padding hidden & \cmark & \xmark & \cmark & 74.38 & 56.09 & 59.67 & 46.61 & 0.616 & \cellcolor{succN}36.69 & \cellcolor{succN}8.20 & \textbf{7.89} & \textbf{0.805} & \textbf{0.505} & \underline{0.936} & \textbf{4.12} & \textbf{4.96} & \underline{15.97} & \underline{0.752} \\
    \rowcolor{oursrow}
    + both (\textbf{\method}) & \cmark & \cmark & \cmark & 74.60 & \underline{56.27} & 59.77 & 47.26 & \textbf{0.619} & \cellcolor{succO}\underline{37.20} & \cellcolor{succO}\underline{10.00} & \underline{7.86} & \textbf{0.805} & \textbf{0.505} & \textbf{0.937} & \underline{4.11} & \textbf{4.96} & \textbf{15.98} & \textbf{0.753} \\
    \bottomrule
  \end{tabular}
\end{table}

\paragraph{Reference assignment steps $\eta_R$.} We apply reference assignment in the first $\eta_R$ of the 50 denoising
steps and leave self-attention unchanged in the remaining steps. \Cref{tab:abl:etar} compares $\eta_R = 10$, 20, 30, 40
and 50. \method is not sensitive to $\eta_R$: \mt{LND} is 36.40,
37.94, 38.00, 36.97 and 37.20 at $\eta_R = 10$, 20, 30, 40 and 50, and \mt{ALL} is 9.20, 9.80, 9.60, 9.40 and 10.00.
No difference from $\eta_R = 50$ exceeds the variation between records: 86, 93, 105 and 88 instructions land only with
$\eta_R = 10$, 20, 30 and 40, against 100, 80, 91 and 92 only with $\eta_R = 50$ (sign test, $p = 0.34$ to 0.82). The other
metrics barely change (\mt{RS} 0.617 to 0.619, \mt{SSO} 0.505 to 0.507). We therefore assign the references in all steps,
which gives the highest \mt{ALL} and leaves no step count to tune.

\begin{table}[!ht]
  \centering
  \caption{\textbf{Reference assignment steps $\eta_R$,} as in \cref{tab:abl:step}. \textbf{Bold} is best, \underline{underlined} second.}
  \label{tab:abl:etar}
  \supptabfont\setlength{\tabcolsep}{3.3pt}\renewcommand{\arraystretch}{0.9}
  \begin{tabular}{@{}l*{15}{c}@{}}
    \toprule
    \multirow{2}{*}{Variant} & \multicolumn{3}{c}{Instr.\ compliance} & \multicolumn{2}{c}{Ref.\ fidelity} & \multicolumn{2}{>{\columncolor{succN}}c}{\textbf{Success}} & \multicolumn{4}{c}{Video fidelity} & \multicolumn{4}{c}{Video quality} \\
    \cmidrule(lr){2-4}\cmidrule(lr){5-6}\cmidrule(lr){7-8}\cmidrule(lr){9-12}\cmidrule(l){13-16}
    & \mt{EXE} & \mt{EQ} & \mt{BND} & \mt{RA} & \mt{RS} & \cellcolor{succN}\textbf{\mt{LND}} & \cellcolor{succN}\textbf{\mt{ALL}} & \mt{SEM} & \mt{SRC} & \mt{SSO} & \mt{MF} & \mt{VQR} & \mt{AES} & \mt{TQ} & \mt{MSM} \\
    \midrule
    $\eta_R = 10$ & 74.49 & 55.92 & 59.58 & 46.80 & \underline{0.618} & \cellcolor{succN}36.40 & \cellcolor{succN}9.20 & 7.84 & \textbf{0.805} & \underline{0.506} & \underline{0.936} & \textbf{4.13} & \textbf{4.97} & 15.92 & \textbf{0.755} \\
    $\eta_R = 20$ & \textbf{74.82} & \textbf{57.37} & \underline{60.15} & 46.76 & \underline{0.618} & \cellcolor{succN}\underline{37.94} & \cellcolor{succN}\underline{9.80} & 7.82 & \textbf{0.805} & \underline{0.506} & \underline{0.936} & \underline{4.12} & \textbf{4.97} & 15.95 & \underline{0.754} \\
    $\eta_R = 30$ & \underline{74.73} & \underline{57.08} & \textbf{60.25} & 46.78 & 0.617 & \cellcolor{succN}\textbf{38.00} & \cellcolor{succN}9.60 & 7.76 & \textbf{0.805} & \textbf{0.507} & \underline{0.936} & \underline{4.12} & \underline{4.96} & 15.96 & 0.753 \\
    $\eta_R = 40$ & 74.65 & 56.47 & 59.58 & \underline{47.05} & \underline{0.618} & \cellcolor{succN}36.97 & \cellcolor{succN}9.40 & \textbf{7.87} & \textbf{0.805} & \underline{0.506} & \textbf{0.937} & \underline{4.12} & \underline{4.96} & \underline{15.97} & 0.753 \\
    \rowcolor{oursrow}
    $\eta_R = 50$ (\textbf{\method}) & 74.60 & 56.27 & 59.77 & \textbf{47.26} & \textbf{0.619} & \cellcolor{succO}37.20 & \cellcolor{succO}\textbf{10.00} & \underline{7.86} & \textbf{0.805} & 0.505 & \textbf{0.937} & 4.11 & \underline{4.96} & \textbf{15.98} & 0.753 \\
    \bottomrule
  \end{tabular}
\end{table}

\paragraph{Instruction assignment blocks.} \Cref{tab:abl:iblocks} applies instruction assignment only in the 10 blocks that
give the masks, in the top 20 blocks, or in all 30 blocks. Assigning in more blocks lands more edits: \mt{LND} rises from 35.37 to 35.54 to 37.20 and \mt{ALL}
from 8.60 to 8.80 to 10.00, as more instructions are executed (\mt{EXE} 71.50 to 74.60) and bound to their targets and
references (\mt{BND} 56.50 to 59.77, \mt{RA} 42.55 to 47.26). Fewer blocks keep slightly more of the untouched video
(\mt{SSO} 0.516 against 0.505), since fewer edits are made. We therefore assign in all blocks.

\begin{table}[!ht]
  \centering
  \caption{\textbf{Instruction assignment blocks,} as in \cref{tab:abl:step}. \textbf{Bold} is best, \underline{underlined} second.}
  \label{tab:abl:iblocks}
  \supptabfont\setlength{\tabcolsep}{3.3pt}\renewcommand{\arraystretch}{0.9}
  \begin{tabular}{@{}l*{15}{c}@{}}
    \toprule
    \multirow{2}{*}{Variant} & \multicolumn{3}{c}{Instr.\ compliance} & \multicolumn{2}{c}{Ref.\ fidelity} & \multicolumn{2}{>{\columncolor{succN}}c}{\textbf{Success}} & \multicolumn{4}{c}{Video fidelity} & \multicolumn{4}{c}{Video quality} \\
    \cmidrule(lr){2-4}\cmidrule(lr){5-6}\cmidrule(lr){7-8}\cmidrule(lr){9-12}\cmidrule(l){13-16}
    & \mt{EXE} & \mt{EQ} & \mt{BND} & \mt{RA} & \mt{RS} & \cellcolor{succN}\textbf{\mt{LND}} & \cellcolor{succN}\textbf{\mt{ALL}} & \mt{SEM} & \mt{SRC} & \mt{SSO} & \mt{MF} & \mt{VQR} & \mt{AES} & \mt{TQ} & \mt{MSM} \\
    \midrule
    The 10 mask blocks & 71.50 & 54.80 & 56.50 & 42.55 & \underline{0.615} & \cellcolor{succN}35.37 & \cellcolor{succN}8.60 & \textbf{7.88} & \textbf{0.807} & \textbf{0.516} & 0.937 & \textbf{4.12} & \underline{4.95} & \underline{15.96} & 0.753 \\
    Top 20 blocks & \underline{73.14} & \underline{55.48} & \underline{58.52} & \underline{46.32} & \underline{0.615} & \cellcolor{succN}\underline{35.54} & \cellcolor{succN}\underline{8.80} & \underline{7.86} & \underline{0.806} & \underline{0.511} & 0.937 & \textbf{4.12} & \underline{4.95} & \textbf{15.98} & 0.753 \\
    \rowcolor{oursrow}
    All 30 blocks (\textbf{\method}) & \textbf{74.60} & \textbf{56.27} & \textbf{59.77} & \textbf{47.26} & \textbf{0.619} & \cellcolor{succO}\textbf{37.20} & \cellcolor{succO}\textbf{10.00} & \underline{7.86} & 0.805 & 0.505 & 0.937 & \underline{4.11} & \textbf{4.96} & \textbf{15.98} & 0.753 \\
    \bottomrule
  \end{tabular}
\end{table}

\paragraph{Instruction assignment steps $\eta_I$.} We apply instruction assignment in the first $\eta_I$ of the 50 denoising
steps and leave cross-attention unchanged in the remaining steps. \Cref{tab:abl:etai} compares $\eta_I = 10$, 20, 30, 40
and 50. Assigning for more steps lands more edits: \mt{LND} rises
from 34.40 at $\eta_I = 10$ to 36.40 at $\eta_I = 20$ and 37.54 at $\eta_I = 40$, with \mt{EXE} from 71.24 to 74.71 and
\mt{RA} from 41.79 to 47.87. $\eta_I = 40$ and $\eta_I = 50$ are close in \mt{LND} (37.54 and 37.20), and $\eta_I = 50$ has the
highest \mt{ALL} (10.00 against 9.20). The video fidelity and quality metrics barely change with $\eta_I$ (\mt{SSO} 0.503 to
0.506). We therefore assign in all steps, which leaves no step count to tune.

\begin{table}[!ht]
  \centering
  \caption{\textbf{Instruction assignment steps $\eta_I$,} as in \cref{tab:abl:step}. \textbf{Bold} is best, \underline{underlined} second.}
  \label{tab:abl:etai}
  \supptabfont\setlength{\tabcolsep}{3.3pt}\renewcommand{\arraystretch}{0.9}
  \begin{tabular}{@{}l*{15}{c}@{}}
    \toprule
    \multirow{2}{*}{Variant} & \multicolumn{3}{c}{Instr.\ compliance} & \multicolumn{2}{c}{Ref.\ fidelity} & \multicolumn{2}{>{\columncolor{succN}}c}{\textbf{Success}} & \multicolumn{4}{c}{Video fidelity} & \multicolumn{4}{c}{Video quality} \\
    \cmidrule(lr){2-4}\cmidrule(lr){5-6}\cmidrule(lr){7-8}\cmidrule(lr){9-12}\cmidrule(l){13-16}
    & \mt{EXE} & \mt{EQ} & \mt{BND} & \mt{RA} & \mt{RS} & \cellcolor{succN}\textbf{\mt{LND}} & \cellcolor{succN}\textbf{\mt{ALL}} & \mt{SEM} & \mt{SRC} & \mt{SSO} & \mt{MF} & \mt{VQR} & \mt{AES} & \mt{TQ} & \mt{MSM} \\
    \midrule
    $\eta_I = 10$ & 71.24 & 53.42 & 55.05 & 41.79 & 0.612 & \cellcolor{succN}34.40 & \cellcolor{succN}8.80 & \underline{7.91} & 0.805 & 0.503 & \textbf{0.937} & \underline{4.11} & 4.96 & \underline{15.95} & 0.753 \\
    $\eta_I = 20$ & 73.49 & 55.70 & 58.13 & 44.82 & 0.616 & \cellcolor{succN}36.40 & \cellcolor{succN}\underline{9.40} & \textbf{7.92} & 0.805 & 0.504 & \textbf{0.937} & \underline{4.11} & 4.96 & \textbf{15.98} & 0.753 \\
    $\eta_I = 30$ & 74.58 & 56.18 & 58.71 & 45.60 & \underline{0.618} & \cellcolor{succN}36.46 & \cellcolor{succN}9.20 & 7.89 & 0.805 & \textbf{0.506} & \underline{0.936} & \textbf{4.12} & 4.96 & \textbf{15.98} & 0.753 \\
    $\eta_I = 40$ & \textbf{74.71} & \textbf{57.22} & \textbf{60.73} & \textbf{47.87} & \underline{0.618} & \cellcolor{succN}\textbf{37.54} & \cellcolor{succN}9.20 & 7.90 & 0.805 & \textbf{0.506} & \textbf{0.937} & \underline{4.11} & 4.96 & \textbf{15.98} & 0.753 \\
    \rowcolor{oursrow}
    $\eta_I = 50$ (\textbf{\method}) & \underline{74.60} & \underline{56.27} & \underline{59.77} & \underline{47.26} & \textbf{0.619} & \cellcolor{succO}\underline{37.20} & \cellcolor{succO}\textbf{10.00} & 7.86 & 0.805 & \underline{0.505} & \textbf{0.937} & \underline{4.11} & 4.96 & \textbf{15.98} & 0.753 \\
    \bottomrule
  \end{tabular}
\end{table}

\paragraph{Instruction assignment against a boost.} Instruction assignment is a hard rule: inside $M_i$, it removes every
other context $c_j$. A softer alternative is a boost, which keeps every context and adds $+\lambda$ to the cross-attention
logits of $c_i$ and $-\lambda$ to the logits of the other $c_j$ inside $M_i$. \Cref{tab:abl:boost} compares assignment with
two boosts at $\lambda = 2$: a flat boost, applied in all 50 steps, and a decaying boost, whose $\lambda$ falls linearly to
0 at the last step. The flat boost matches assignment. Its \mt{LND} is 37.03 against 37.20 and \mt{ALL} 9.60 against
10.00, and about as many instructions land only with the flat boost (104) as only with assignment (107). The \mt{EXE} and
\mt{BND} of the flat boost are slightly higher (75.95 and 61.50 against 74.60 and 59.77). The match is expected: at
$\lambda = 2$, the flat boost already scales the attention to each other $c_j$ by $e^{-2\lambda} \approx 0.02$ relative to
$c_i$, and assignment is the limit of the boost as $\lambda \rightarrow \infty$. The decaying boost lands fewer edits
(\mt{LND} 36.23, \mt{ALL} 7.40), since the other contexts return in the late steps. Assignment thus reaches the result
of the best boost with no strength or schedule to tune.

\begin{table}[!ht]
  \centering
  \caption{\textbf{Instruction assignment against a boost,} as in \cref{tab:abl:step}. \textbf{Bold} is best, \underline{underlined} second.}
  \label{tab:abl:boost}
  \supptabfont\setlength{\tabcolsep}{3.3pt}\renewcommand{\arraystretch}{0.9}
  \begin{tabular}{@{}l*{15}{c}@{}}
    \toprule
    \multirow{2}{*}{Variant} & \multicolumn{3}{c}{Instr.\ compliance} & \multicolumn{2}{c}{Ref.\ fidelity} & \multicolumn{2}{>{\columncolor{succN}}c}{\textbf{Success}} & \multicolumn{4}{c}{Video fidelity} & \multicolumn{4}{c}{Video quality} \\
    \cmidrule(lr){2-4}\cmidrule(lr){5-6}\cmidrule(lr){7-8}\cmidrule(lr){9-12}\cmidrule(l){13-16}
    & \mt{EXE} & \mt{EQ} & \mt{BND} & \mt{RA} & \mt{RS} & \cellcolor{succN}\textbf{\mt{LND}} & \cellcolor{succN}\textbf{\mt{ALL}} & \mt{SEM} & \mt{SRC} & \mt{SSO} & \mt{MF} & \mt{VQR} & \mt{AES} & \mt{TQ} & \mt{MSM} \\
    \midrule
    Boost, $\lambda = 2$ & \textbf{75.95} & \textbf{57.92} & \textbf{61.50} & \textbf{47.81} & \underline{0.619} & \cellcolor{succN}\underline{37.03} & \cellcolor{succN}\underline{9.60} & 7.82 & \textbf{0.806} & \underline{0.508} & \textbf{0.937} & \textbf{4.14} & \underline{4.95} & \textbf{16.08} & \textbf{0.754} \\
    Boost, $\lambda = 2$, linear decay to 0 & \underline{74.86} & \underline{57.39} & \underline{60.54} & 46.95 & \textbf{0.620} & \cellcolor{succN}36.23 & \cellcolor{succN}7.40 & \underline{7.85} & \textbf{0.806} & \textbf{0.509} & \underline{0.936} & \textbf{4.14} & \underline{4.95} & \underline{16.07} & \textbf{0.754} \\
    \rowcolor{oursrow}
    Assignment (\textbf{\method}) & 74.60 & 56.27 & 59.77 & \underline{47.26} & \underline{0.619} & \cellcolor{succO}\textbf{37.20} & \cellcolor{succO}\textbf{10.00} & \textbf{7.86} & \underline{0.805} & 0.505 & \textbf{0.937} & \underline{4.11} & \textbf{4.96} & 15.98 & \underline{0.753} \\
    \bottomrule
  \end{tabular}
\end{table}

\subsection{\texorpdfstring{\method}{TACO-Zero} on another public benchmark}
\label{sec:app:textonly}
\label{sec:exp:textonly}
\bench is our own benchmark. This section tests \method on UniVBench~\cite{univbench}, a public benchmark with its own
evaluator, UniV-Eval.

\paragraph{Setup.} A UniVBench record has one compound instruction that names its references by number. The joint prompt
takes this instruction and all references as they are. For \method, we split the instruction into its clauses, each with at
most one reference, and keep a record only when the clauses rejoin to the original instruction word for word. This gives 438 records with two or more instructions and at least one reference. UniV-Eval rates
each result on eight dimensions and averages them. Its own judge is closed, so we run UniV-Eval unchanged
with Qwen3.5-122B-A10B~\cite{qwen35}, the judge of \bench.

\paragraph{Results.} On Kiwi-Edit, \method raises the average from 46.3 to 49.1 and improves all eight dimensions
(\cref{tab:textonly}). On JoyAI-Video-Edit, the average is 57.1 with the joint prompt and 55.9 with \method, a paired
difference of $-1.2 \pm 1.1$, so \method does not improve JoyAI-Video-Edit on this benchmark.

\begin{table}[!ht]
  \centering
  \caption{\textbf{\method on UniVBench}~\cite{univbench}. UniV-Eval's dimensions and their average, with Qwen3.5-122B-A10B~\cite{qwen35} as
  its judge, on the 438 records with two or more instructions and at least one reference. In each pair, \textbf{bold} is better.}
  \label{tab:textonly}
  \supptabfont\setlength{\tabcolsep}{4.2pt}
  \begin{tabular}{@{}l*{9}{c}}
    \toprule
    \multirow{2}{*}{Method} & \multicolumn{9}{c}{UniV-Eval} \\
    \cmidrule(l){2-10}
    & Subject & Backgr. & Action & Camera & Color & Light. & Style & Rel.\ pos. & \cellcolor{succN}\textbf{Avg.} \\
    \midrule
    \rowcolor{baserow}
    Kiwi-Edit (KE)~\cite{kiwiedit}, joint prompt & 20.5 & 36.8 & 39.4 & 70.5 & 50.6 & 61.6 & 49.6 & 41.6 & \cellcolor{succB}46.3 \\
    \rowcolor{oursrow}
    KE + \textbf{\method} (Ours) & \textbf{23.2} & \textbf{40.6} & \textbf{42.8} & \textbf{72.4} & \textbf{53.0} & \textbf{63.6} & \textbf{50.9} & \textbf{46.5} & \cellcolor{succO}\textbf{49.1} \\
    \midrule
    \rowcolor{baserow}
    JoyAI (JAI)~\cite{joyai}, joint prompt & \textbf{34.8} & \textbf{52.2} & \textbf{53.3} & \textbf{78.6} & \textbf{59.1} & \textbf{68.4} & \textbf{53.9} & 56.2 & \cellcolor{succB}\textbf{57.1} \\
    \rowcolor{oursrow}
    JAI + \textbf{\method} (Ours) & 33.2 & 51.4 & 51.4 & 77.4 & 58.5 & 67.9 & 50.8 & \textbf{56.4} & \cellcolor{succO}55.9 \\
    \bottomrule
  \end{tabular}
\end{table}

\paragraph{By record type.} \Cref{tab:textonly:split} splits the records into two types. In 312 records, every instruction
has a reference. In the other 126 records, at least one instruction has no reference. On the first type, \method raises
Kiwi-Edit by $4.0 \pm 1.2$ and improves all eight dimensions, most on Relative position ($+6.6$), Background ($+6.4$), Action
($+4.8$) and Subject ($+4.3$). JoyAI-Video-Edit is unchanged ($-0.3 \pm 1.4$), and none of its dimensions moves by more
than about one standard error. On the second type, Kiwi-Edit is unchanged ($-0.1 \pm 1.9$) and JoyAI-Video-Edit loses
$3.4 \pm 1.9$, mostly on Subject ($-8.2 \pm 2.6$). UniV-Eval ties no score to an instruction, so it cannot tell which
instructions cause this loss. \bench scores each instruction on its own, and there \method does not lower the instructions
without a reference: over these 628 instructions, \mt{LND} rises from 30.25 to 42.36 on Kiwi-Edit and from 52.71 to 54.78
on JoyAI-Video-Edit. We leave the loss on the 126 UniVBench records to future work.

\paragraph{By number of references.} UniVBench has few records with more than two references: of the 312 records, 289 have
two and 23 have three. With two references, the joint prompt of JoyAI-Video-Edit scores 58.6 and \method changes little
($-0.6 \pm 1.5$). With three, the joint prompt falls to 50.0 and \method reaches 54.0 ($+4.0 \pm 3.8$), and Kiwi-Edit gains
$3.8 \pm 4.2$. These 23 records are too few for a firm result, but the trend agrees with \bench, where the joint prompt
lands fewer edits as references are added (\cref{sec:exp}). UniVBench therefore tests mostly the two-reference case, where
JoyAI-Video-Edit is already strong, while \bench balances $n$ from one to five.

\begin{table}[!ht]
  \centering
  \caption{\textbf{UniVBench by record type.} UniV-Eval average of \cref{tab:textonly}, split by whether every instruction of a record has a
  reference, and by the number of references. $\Delta$: \method minus the joint prompt, with the standard error of the paired difference over records.}
  \label{tab:textonly:split}
  \supptabfont\setlength{\tabcolsep}{5pt}
  \begin{tabular}{@{}lc*{6}{c}@{}}
    \toprule
    \multirow{2}{*}{Records} & \multirow{2}{*}{\#} & \multicolumn{3}{c}{Kiwi-Edit (KE)~\cite{kiwiedit}} & \multicolumn{3}{c}{JoyAI (JAI)~\cite{joyai}} \\
    \cmidrule(lr){3-5}\cmidrule(l){6-8}
    & & Joint prompt & + \textbf{\method} & $\Delta$ & Joint prompt & + \textbf{\method} & $\Delta$ \\
    \midrule
    Every instruction has a reference & 312 & 46.0 & \textbf{50.0} & $+4.0 \pm 1.2$ & \textbf{58.0} & 57.7 & $-0.3 \pm 1.4$ \\
    \quad with two references & 289 & 46.1 & \textbf{50.1} & $+4.0 \pm 1.3$ & \textbf{58.6} & 58.0 & $-0.6 \pm 1.5$ \\
    \quad with three references & 23 & 44.8 & \textbf{48.6} & $+3.8 \pm 4.2$ & 50.0 & \textbf{54.0} & $+4.0 \pm 3.8$ \\
    Some instruction has no reference & 126 & \textbf{47.1} & 47.0 & $-0.1 \pm 1.9$ & \textbf{54.9} & 51.4 & $-3.4 \pm 1.9$ \\
    \midrule
    All & 438 & 46.3 & \textbf{49.1} & $+2.8 \pm 1.0$ & \textbf{57.1} & 55.9 & $-1.2 \pm 1.1$ \\
    \bottomrule
  \end{tabular}
\end{table}

\subsection{Runtime and memory}
\label{sec:app:cost}
\Cref{tab:cost} reports the runtime per record and the peak GPU memory of every method that runs on our GPUs, each on
one NVIDIA B200, as medians over the same 80 records, 20 for each $k$.

\paragraph{Runtime.} On Kiwi-Edit, \method takes 39 to 53 seconds per record, against 18 to 20 for the joint prompt and
37 to 93 for sequential editing: each \textit{Locate} pass stops after a tenth of the steps, so the cost grows slowly with $k$. On the
few-step JoyAI-Video-Edit, a \textit{Locate} pass is a full render, so \textit{Locate} adds more (111 to 276 seconds, against 32 to 36 for the joint prompt and
63 to 156 for sequential editing). Kiwi-Edit + \method is faster than every baseline except Kiwi-Edit-Ref (18 to 20 seconds).
JoyAI-Video-Edit + \method is slower than LoomVideo (60 to 64) and LTX-Ripple (101 to 103) and faster than the rest,
which take 347 to 5{,}201 seconds.

\paragraph{Memory.} \method adds almost no memory, since the \textit{Locate} passes run one instruction at a time: on Kiwi-Edit, the peak
is 22.9 to 23.1\,GiB against 22.7 for the joint prompt, and on JoyAI-Video-Edit it is 21.5 to 23.4\,GiB against 23.9 to 26.7.
The other baselines need from 15.0\,GiB (ReCo) to 92.9\,GiB (MiniMax-H3).

\begin{table}[!ht]
  \centering
  \caption{\textbf{Runtime and peak memory by $k$} on \bench: the median over the records of each $k$ of the runtime per
  record and of the peak allocated GPU memory, each method on one NVIDIA B200. Seedance~2.5~\cite{seedance25} runs behind an API and is
  not listed.}
  \label{tab:cost}
  \supptabfont\setlength{\tabcolsep}{6pt}
  \begin{tabular}{@{}l*{8}{c}@{}}
    \toprule
    \multirow{2}{*}{Method} & \multicolumn{4}{c}{Runtime (s), by $k$ $\downarrow$} & \multicolumn{4}{c}{Peak memory (GiB), by $k$ $\downarrow$} \\
    \cmidrule(lr){2-5}\cmidrule(l){6-9}
    & 2 & 3 & 4 & 5 & 2 & 3 & 4 & 5 \\
    \midrule
    CoinVE-Edit~\cite{coinve} & 587 & 605 & 642 & 652 & 50.5 & 50.5 & 50.5 & 50.6 \\
    Bernini~\cite{bernini} & 4{,}326 & 5{,}201 & 4{,}464 & 3{,}478 & 51.6 & 51.9 & 52.6 & 52.9 \\
    MiniMax-H3~\cite{minimaxh3} & 2{,}291 & 2{,}396 & 2{,}706 & 2{,}779 & 84.0 & 88.1 & 88.2 & 92.9 \\
    LoomVideo~\cite{loomvideo} & 59.5 & 63.0 & 63.8 & 63.8 & 41.1 & 41.2 & 41.5 & 41.7 \\
    Kiwi-Edit-Ref~\cite{kiwiedit} & 17.9 & 18.4 & 19.6 & 19.4 & 22.7 & 22.7 & 22.7 & 22.7 \\
    ReCo~\cite{reco} & 2{,}743 & 4{,}153 & 5{,}187 & 4{,}610 & 15.0 & 15.5 & 15.5 & 15.5 \\
    LTX-Ripple~\cite{ltxripple} & 101 & 102 & 103 & 103 & 76.4 & 76.4 & 76.4 & 76.4 \\
    EditVid~\cite{editvid} & 347 & 387 & 430 & 474 & 56.6 & 64.3 & 71.9 & 79.7 \\
    \midrule
    Kiwi-Edit (KE)~\cite{kiwiedit}, sequential & 37.1 & 54.3 & 74.5 & 93.0 & 22.7 & 22.7 & 22.7 & 22.7 \\
    \rowcolor{baserow}
    KE, joint prompt & 18.5 & 18.1 & 18.8 & 20.1 & 22.7 & 22.7 & 22.7 & 22.7 \\
    \rowcolor{oursrow}
    KE + \textbf{\method} (Ours) & 39.0 & 42.9 & 49.2 & 53.0 & 22.9 & 23.0 & 23.0 & 23.1 \\
    \midrule
    JoyAI (JAI)~\cite{joyai}, sequential & 62.5 & 89.3 & 125 & 156 & 23.0 & 23.0 & 23.0 & 23.0 \\
    \rowcolor{baserow}
    JAI, joint prompt & 31.9 & 32.2 & 35.3 & 35.5 & 23.9 & 24.8 & 25.8 & 26.7 \\
    \rowcolor{oursrow}
    JAI + \textbf{\method} (Ours) & 111 & 149 & 215 & 276 & 21.5 & 22.1 & 22.8 & 23.4 \\
    \bottomrule
  \end{tabular}
\end{table}

\suppnobreak
\subsection{More qualitative results}
\label{sec:app:qual}
\Cref{fig:app:qual:port,fig:app:qual} show six more records, three portrait and three landscape, each set in order of growing
difficulty. As in \cref{fig:qual}, a red arrow marks an edit that a backbone misses, a green arrow the same edit landed by \method
on that backbone, and an orange arrow a partial edit or an unwanted change.

\paragraph{Portrait records (\cref{fig:app:qual:port}).} On the cyclist ($k = 3$), Kiwi-Edit keeps the grey shirt, and \method gives the orange jersey and
lands all three edits. JoyAI-Video-Edit keeps the white shoes, and \method turns them orange. At the coffee press ($k = 4$),
Kiwi-Edit keeps the white tray, and \method swaps in the wooden board. JoyAI-Video-Edit adds no coaster, and \method adds it and
lands all four edits. In the caf\'e ($k = 5$), Kiwi-Edit keeps the marble table and the green mug, and \method gives the walnut table
and the burgundy mug. JoyAI-Video-Edit adds the vase in front of the glass of water, not beside it, while \method places the vase
beside the glass.

\paragraph{Landscape records (\cref{fig:app:qual}).} On the bird ($k = 3$), Kiwi-Edit drops the wooden dowel, and JoyAI-Video-Edit leaves a pale branch
across the new seashore. With \method, Kiwi-Edit keeps the bird on the dowel, and JoyAI-Video-Edit shows a clean sea behind the bird.
The baselines change other parts of the scene: Seedance~2.5 removes the metal bowl, Bernini adds a second bird, LoomVideo paints part
of the dowel green, and LTX-Ripple adds a wall. On the sleeping dog ($k = 3$, the record of \cref{fig:ablation}), Kiwi-Edit adds only
the basket, and \method adds the mug and the frisbee as well. JoyAI-Video-Edit doubles the basket, and \method gives a single one.
Bernini places the frisbee away from the paws, and LTX-Ripple replaces the lawn behind the dog with a wall. In the living room
($k = 5$), Kiwi-Edit adds no fern and keeps the brown armchair, and \method adds the fern and the burgundy armchair. \method lands all
five edits on both backbones, while LoomVideo replaces the whole scene.

\suppsec{Limitations}
\label{sec:app:limits}

\label{sec:app:dependent}\label{sec:app:otherlimits}%
\paragraph{Several edits on one object.}
\method assumes that the targets of the instructions are disjoint. \textit{Locate} gives each video token to one instruction
(\cref{eq:mask}), so instructions that edit the same object compete for its tokens. In \mbox{\cref{fig:app:fail}-(A)}, four
instructions edit one dog and a fifth adds a rug under it. With \method, the bandana, the sunglasses, the backpack and the rug
land on both backbones, but the dog stays brown: the regions of $e_2$ to $e_5$ take the dog's face, neck, back and outline, and
the recolor $e_1$ keeps only a part of its body. With the joint prompt, JoyAI-Video-Edit~\cite{joyai} lands all five edits and
Kiwi-Edit~\cite{kiwiedit} darkens most of the dog. \method is therefore worse than the joint prompt in this case. A token would
have to follow several instructions, and letting the masks overlap does not land more edits on \bench (\cref{tab:abl:overlap}).
A possible remedy is to group the instructions whose candidate sets $\tilde{M}_i$ overlap heavily: the group is encoded together
as one context, and this context gets the union of the group's regions. The shared object would then be edited as under the
joint prompt, which handles this case better, while the other instructions stay separate. We leave this grouping to future
work.

\paragraph{Tasks the backbone does not support.}
\method changes no weights, so it composes only the edits that its backbone can make. In \mbox{\cref{fig:app:fail}-(B)}, $e_1$
writes two words on a whiteboard, an edit that no instruction of \bench asks for. Kiwi-Edit draws unreadable marks with the
joint prompt and with \method, although \textit{Locate} finds the whiteboard and the other two edits land.
JoyAI-Video-Edit writes the words with the joint prompt, and \method keeps them. \method thus neither adds an ability to its
backbone nor removes one. A remedy is to teach the backbone the missing task
alone. The backbone can be fine-tuned on a single-instruction dataset, here a scene text editing
dataset, and \method then composes the new edit with the other edits zero-shot at inference, so no
compositional training data is needed. \method is also not tied to one backbone: on JoyAI-Video-Edit, which already writes
text, the same record gets its words.

\clearpage
\vspace*{\fill}%
\begin{figure}[!t]%
  \centering
  \includegraphics[width=\textwidth]{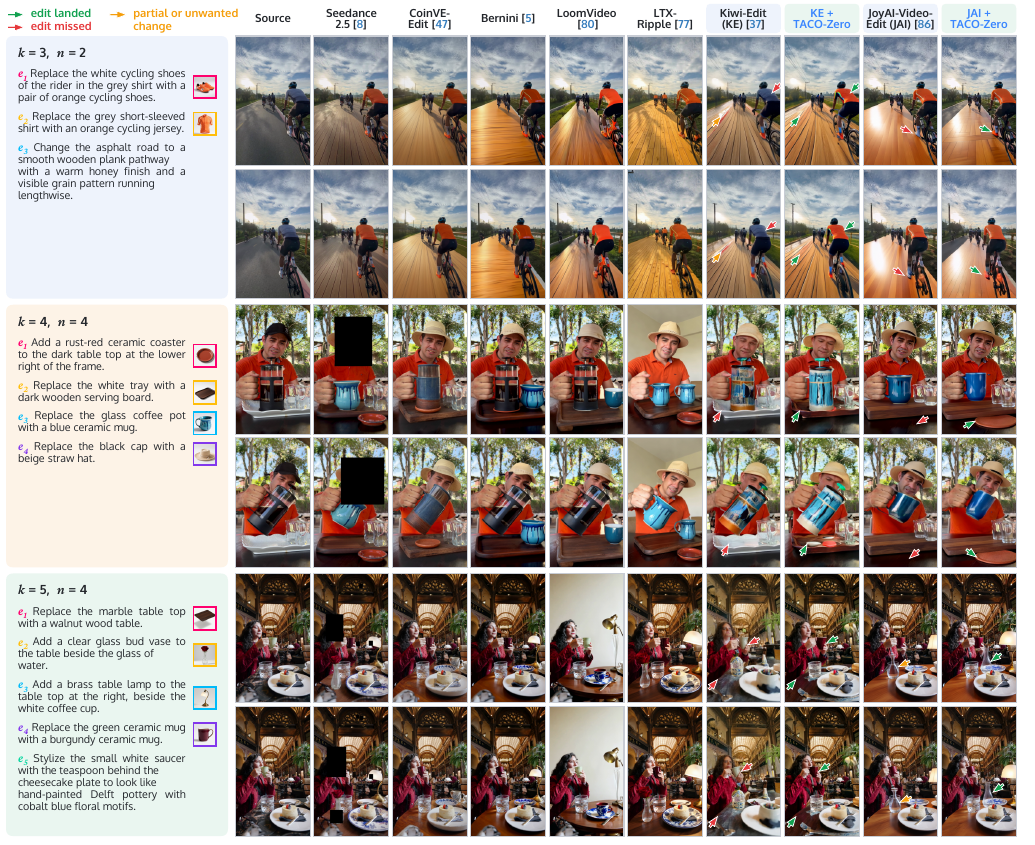}
  \vspace{-20pt}%
  \caption{\textbf{More qualitative results on portrait records} ($k = 3$, 4, 5), arrows as in \cref{fig:qual}. Videos are on our \webpage.}
  \label{fig:app:qual:port}
\end{figure}
\vspace*{\fill}%

\clearpage
\vspace*{\fill}%
\begin{figure}[!t]%
  \centering
  \includegraphics[width=\textwidth]{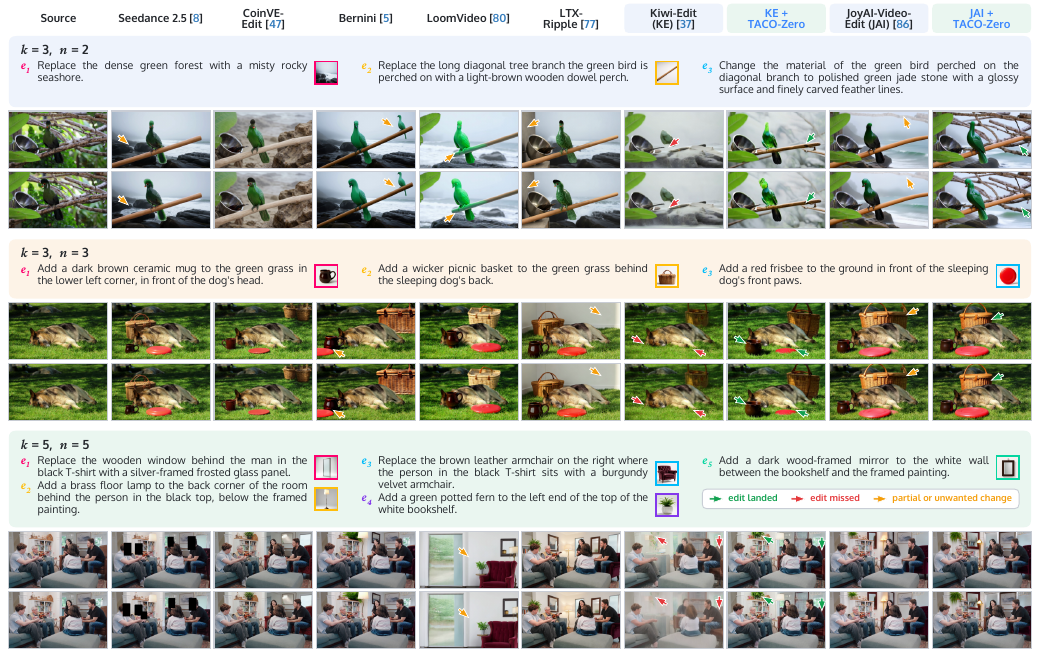}
  \vspace{-20pt}%
  \caption{\textbf{More qualitative results on landscape records} ($k = 3$, 3, 5), arrows as in \cref{fig:qual}. Videos are on our \webpage.}
  \label{fig:app:qual}
\end{figure}
\vspace*{\fill}%

\clearpage
\vspace*{\fill}%
\begin{figure}[!ht]
  \centering
  \includegraphics[width=\textwidth,height=0.85\textheight,keepaspectratio]{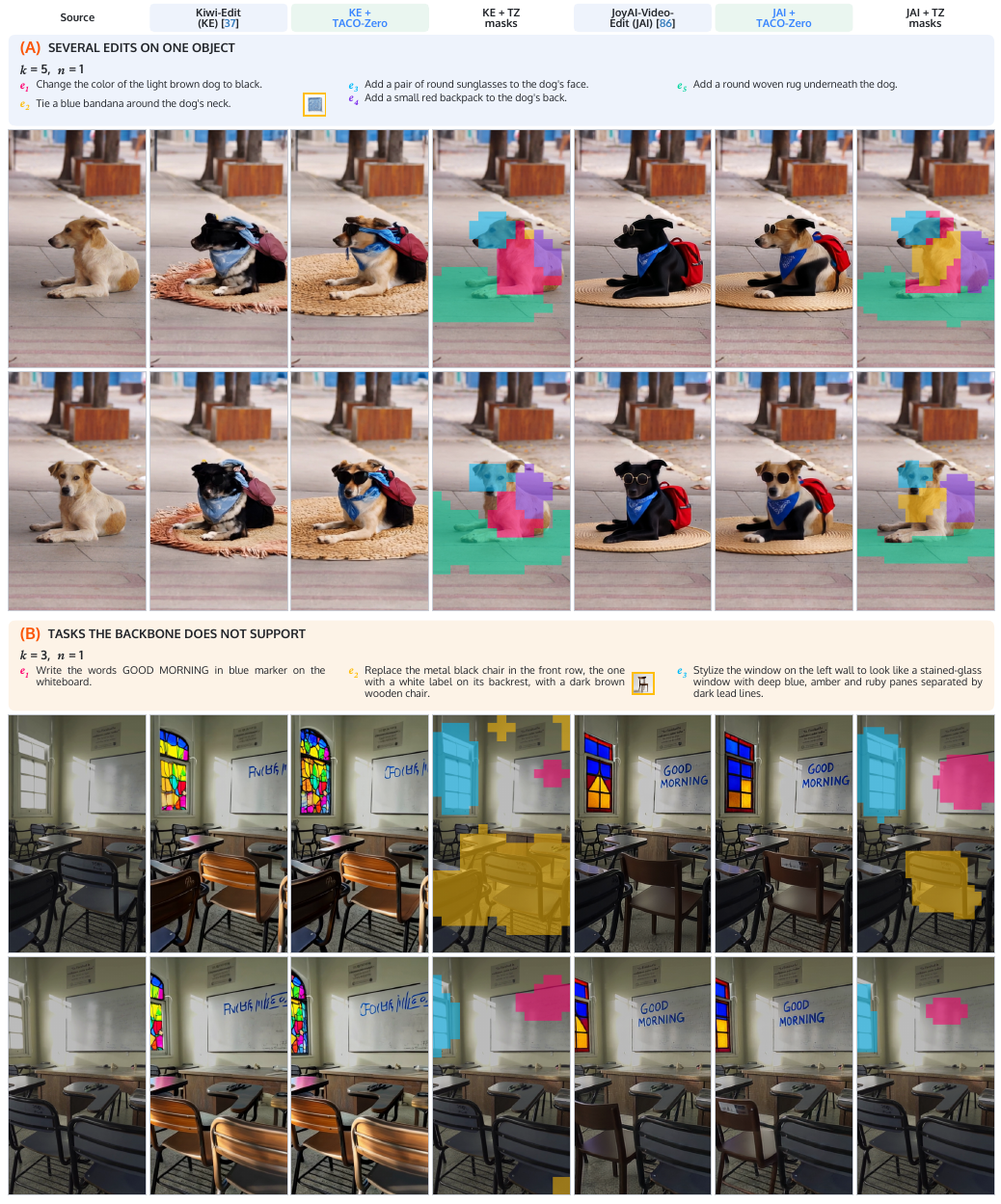}
  \vspace{-6pt}%
  \caption{\textbf{Limitations.} (A) Several edits on one object: with \method, the dog stays brown on both backbones, because
  the regions of $e_2$ to $e_5$ take most of its tokens. (B) Tasks the backbone does not support: Kiwi-Edit cannot write the
  words, with or without \method, while JoyAI-Video-Edit can. Each case shows two frames. The mask columns tint the region $M_i$
  of each instruction in the color of $e_i$ on the source frame.}
  \label{fig:app:fail}
\end{figure}
\vspace*{\fill}%
\clearpage

\fi

\end{document}